\documentclass[conference]{IEEEtran}
\usepackage{times}

\usepackage[numbers]{natbib}
\usepackage{multicol}
\usepackage[bookmarks=true]{hyperref}
\usepackage{orcidlink}

\usepackage{graphicx}%
\usepackage{multirow}%
\usepackage{amsmath,amssymb,amsfonts}%
\usepackage{amsthm}%
\usepackage{mathrsfs}%
\usepackage{pifont}

\usepackage{tabularx}
\usepackage{xcolor}%
\usepackage{textcomp}%
\usepackage{manyfoot}%
\usepackage{booktabs}%
\usepackage{algorithm}%
\usepackage{algorithmicx}%
\usepackage{algpseudocode}%
\usepackage{listings}%

\usepackage{orcidlink}%

\usepackage{xurl}
\usepackage{placeins}
\usepackage{subfigure}
\usepackage{dblfloatfix}
\usepackage{acronym}
\usepackage{siunitx}
\usepackage{bm}
\usepackage{svg}
\usepackage{newtxmath} 
\usepackage[scr=rsfso]{mathalfa} 
\usepackage[noabbrev, capitalise]{cleveref}
\usepackage[perpage]{footmisc}

\acrodef{RL}{Reinforcement Learning}
\acrodef{NN}{Neural Network}
\acrodef{DoF}{Degree of Freedom}
\acrodef{ID}{Inverse Dynamics}
\acrodef{IMU}{Inertial Measurement Unit}
\acrodef{PPO}{Proximal Policy Optimization}
\acrodef{PSD}{Power Spectral Density}
\acrodef{MLP}{Multi-Layer Perceptron}
\acrodef{COG}{Center of Gravity}
\acrodef{EE}{end-effector}
\acrodef{IK}{Inverse Kinematics}
\acrodef{AOA}{angle of arrival}
\acrodef{RF}{radio frequency}
\acrodef{SDR}{Software Defined Radio}
\acrodef{SM}{Switching Matrix}
\acrodef{CW}{Continuous Wave}
\acrodef{ISM}{Industrial, Scientific, and Medical}
\acrodef{TDM}{Time Division Multiplexing}
\acrodef{NFC}{Negative Flow Control}
\acrodef{PFC}{Positive Flow Control}
\acrodef{LS}{Load Sensing}
\acrodef{MPC}{Model Predictive Control}
\acrodef{FF}{Feed Forward}
\acrodef{ECU}{Engine Control Unit}
\acrodef{DCV}{Directional Control Valve}
\acrodef{LUT}{Lookup Table}
\acrodef{MISO}{Multiple Input Single Output}

\begin{document}

\title{High Precision Hydraulic Excavator Control for Heavy-Duty Grading}

\author{
\IEEEauthorblockN{Lennart Werner$^{1}$, Pol Eyschen$^{1}$, Sean Costello$^{2}$, Andrei Cramariuc$^{1}$ and Marco Hutter$^{1}$}
\IEEEauthorblockA{$^{1}$ETH Zürich, Robotic Systems Lab, Zürich, Switzerland}
\IEEEauthorblockA{$^{2}$Hexagon AB, Heerbrugg, Switzerland}
\IEEEauthorblockA{Email: \{lennartwerner, peyschen, crandrei, mahutter\}@ethz.ch, sean.costello@hexagon.com}
}

\maketitle
\begin{abstract}
High-precision heavy-duty grading is a common step in earthworks, traditionally carried out manually by skilled operators.
Removing a significant amount of material while achieving a high-precision surface requires substantial machine-specific experience.
Different hydraulic architectures react differently to operator inputs and soil interaction forces, which makes generalizable controllers challenging.
In this paper, we present an autonomous controller that achieves high-precision grading at expert-operator speed on Load Sensing and Negative Flow Control machines alike.
We split our controller into two parts: (1) a hydraulic-aware low-level loop that is hydraulic architecture-specific and (2) a path-tracking layer that coordinates joint motions and responses.
Through a calibration process, our technique is applicable to load-sensing and negative-flow-control machinery.
To showcase its versatility, we benchmark our approach on two excavators with different hydraulics and compare it against a commercial state-of-the-art solution.
Our technique (RMSE 1.8~cm) outperforms the commercial solution (RMSE 4.7~cm) in precision by a factor of 2.6 and improves machine usage by leveraging the maximum function pressure, as opposed to commercial solutions that stall prematurely.
\end{abstract}

\section{Introduction}\label{sec1}
Despite significant progress in industrial automation, heavy machinery on construction sites is still largely operated manually. 
Operators are exposed to hazardous environments and face high physical and cognitive loads. 
At the same time, the industry struggles to meet demand due to persistent labor shortages. 
Intelligent assistive systems and automation for heavy equipment can enable safer, more consistent, and more efficient operation, especially for users with limited experience. 
Yet automating earthmoving tasks such as digging, trenching, and grading is difficult due to the complex coupling between the soil and the machine as well as the high complexity associated with precision hydraulic control.
In particular, high-precision tasks such as grading require substantial operator experience and are time-consuming to perform.

\begin{figure}
    \centering
    \includegraphics[width=\linewidth]{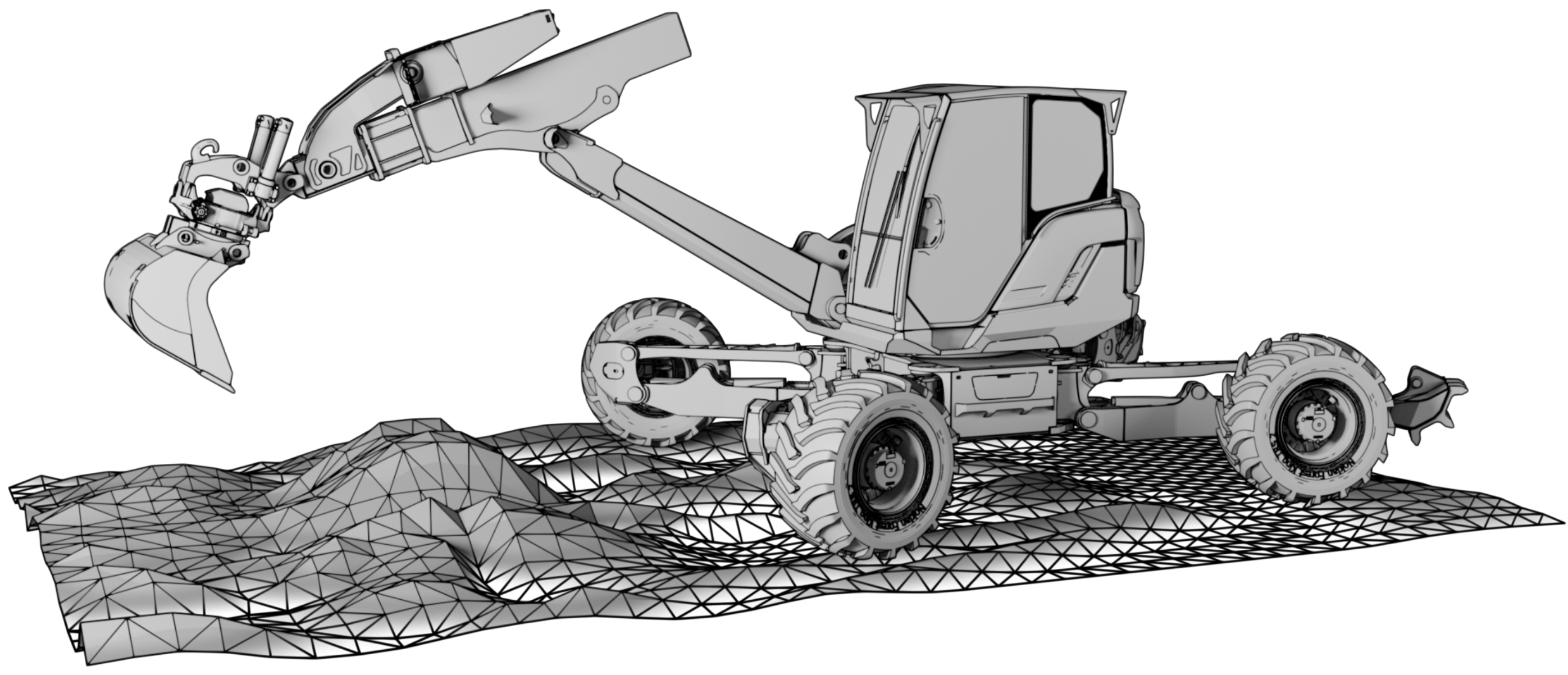}
    \vspace{-20pt}
    \caption{Heavy-duty grading with the Menzi Muck M445. Grading denotes the removal of uneven ground to form a prescribed planar surface. In the shown setting, this requires substantial soil interaction while maintaining the bucket edge on the target plane. The proposed retrofittable, hydraulics-aware controller automates such in-soil grading passes across excavators and achieves centimeter-level surface accuracy.}
    \label{fig:grading_fig1}
\end{figure}

In this paper, we address autonomous high-accuracy grading with substantial ground interaction, as shown in Figure~\ref{fig:grading_fig1} and in our supplementary video\footnote{\href{video}{https://youtu.be/bCOMYbRWv5I}}.
We propose a technique that generalizes across excavators and hydraulic architectures such as \ac{LS} and \ac{NFC}. 
We define heavy-duty grading as cuts that fill the bucket in less than one stroke.
Current commercial systems perform poorly in deep grading, stall before the maximum function pressure is reached, and deviate from the target surface. 
State-of-the-art \ac{IK}-based methods also lose accuracy across soils, since joint responses are load-dependent and are not fully modeled~\cite{Yang_Zhang_Hong_Chen_Yang_Wang_Cao_2022}.
Prior work has not shown real in-soil centimeter-level grading that combines soft hydraulic actuation, load-dependent compensation, short calibration, and transfer across hydraulic architectures.

We propose a retrofittable\footnote{can be calibrated without vendor-specific access}, machine-agnostic\footnote{generalizes across machines without changes to the method} controller for \ac{LS} and \ac{NFC} hydraulic excavators.  
It combines \ac{NFC}-specific hydraulic compensation for soft, load-dependent actuation, a fast one-time calibration without vendor-specific access, and a \ac{MPC} path-tracking layer for precise in-soil grading under varying load. 
\ac{LS} hydraulics are modeled with a simple mapping between \ac{DCV} displacement and function flow, while \ac{NFC} machines require a load-dependent function flow model. 
We introduce a new \ac{NFC} model that accurately inverts the hydraulically soft behavior and matches the target function flow under load. 
Velocity transients are modeled as a second-order system with added dead time for the \ac{MPC} path tracking.

We validate the method on two excavators with different weight classes and hydraulics. 
Velocity tracking experiments show that the calibrated hydraulic models match target joint rates under varying loads.  
We compare in-soil deep grading performance to a commercial state-of-the-art system. 
In a test campaign on the \ac{NFC} excavator, our controller achieves an RMSE height error of \SI{1.8}{\cm} at different cutting depths, compared to \SI{4.7}{\cm} for the commercial system. 
Experiments on the \SI{11.5}{\tonne} \ac{LS} machine confirm a precision of \SI{1.4}{\cm} across various cutting depths. 
We assess surface quality and precision of the graded areas with a high-precision laser scanner. 
The proposed system stalls only at the maximum function pressure, unlike commercial controllers, which stall earlier, thereby under-utilizing the machinery. Our framework presents a significant advance in grading quality, enabling direct use of the prepared terrain.
In our tested setting, the commercial system required additional passes to reach comparable surface quality.

In sum, the contributions of this paper are:
\begin{itemize}
    \item A hydraulics-aware control stack for heavy-duty high-precision grading that achieves 2.6 times lower error than the baseline.
    \item A novel pressure-adaptive \ac{NFC} model that compensates soft, load-dependent hydraulic actuation.
    \item A one-time retrofit calibration routine and path-tracking layer that achieve sub-\SI{2}{\cm} in-soil grading.
\end{itemize}

\section{Related Work}
\label{sec:related_work}
Automation of earthmoving equipment has been investigated from early assistive systems to fully autonomous platforms. 
Reviews of earthmoving automation and construction robotics summarize sensing, perception, and control approaches across excavators, wheel loaders, and bulldozers, and highlight the trend toward tighter integration of machine control with digital site models and fleet-level planning~\cite{Azar_Kamat,Nguyen_Ha_2023}.

\subsection{Automation of Heavy Machinery}
Robotic excavation has progressed from early research prototypes to full-scale autonomous machines.
Free-form trenching and foundation work is demonstrated using a walking excavator in~\cite{Jud19AutonomousFreeForm}. 
Autonomous short-cycle loading with standard tracked excavators has been shown in a complete system that performs material loading in a quarry, including perception, planning, and robust execution in varied terrain~\cite{Zhang_Zhao_Long_Wang_Qian_Lu_Song_Manocha_2021}. 
Recent work on excavation planning extends these results to higher-level task and motion planning for bulk earth removal~\cite{Terenzi_Hutter_2024}.

Beyond excavators, autonomous and semi-autonomous wheel loaders and haul trucks have been studied for mining and earthworks. Simulation-based optimization and, more recently, learned world models are used to tune loading cycles and bucket trajectories under granular material interaction~\cite{Aoshima_Servin_2024}. These systems focus on productivity and cycle time and typically rely on proprietary machine-control layers tuned to a specific hydraulic architecture and task.

\subsection{Hydraulic Motion Control}
Hydraulic excavator motion control research spans physics-based modeling, optimization, and learned abstractions. 
A complete grading pipeline with perception, stroke planning, and motion control is presented by \citet{10697977}. 
On the modeling side, \citet{Kim_Kim_Kim_Lee_2019} derive a nonlinear \ac{DCV} model capturing internal switching, joint coupling, and shared pump-flow constraints, and use constrained optimization to scale the desired bucket motion under flow saturation.

Learning-based and reduced-order approaches often trade modeling effort for data or structural assumptions. 
\citet{Lee_Choi_Kim_Moon_Kim_Lee_2022} learn an inverse mapping from desired joint velocities to valve commands with delay compensation, reporting $\approx \SI{2}{\cm}$ tracking after several hours of machine-specific data collection.
\citet{Wang_Zhang_Hao_Deng_2023} propose an observer-based approximate affine nonlinear \ac{MPC} to improve tracking with reduced computational cost, and \citet{Msaad_Cecchin_Demir_Fagiano_2025} apply data-driven nonlinear \ac{MPC} via local model networks for grading on a single machine and hydraulic architecture.

Reinforcement-learning methods have been tested on full-scale hydraulic machines, including sim-to-real trajectory control~\cite{Egli_Hutter_2022}, end-to-end bucket filling for soil manipulation~\cite{10616135}, offline \ac{RL} for rigid-object excavation~\cite{Jin_Ye_Zhang_2023,Lu_Zhu_Zhang_2022}, and learned rock or bulk manipulation~\cite{Gruetter_Terenzi_Egli_Hutter_2025}. 
For grading, \citet{Kim_Kim_2025} use \ac{RL} to tune a neural-network inversion layer, while \citet{Spinelli_Egli_Nubert_Nan_Bleumer_Goegler_Brockes_Hofmann_Hutter_2024} and \citet{10802743} control hydraulic material-handling machines with underactuated tools.
In our preliminary light-grading tests, an end-to-end \ac{RL} policy reached about \SI{6}{\cm} tracking error.
It relied on a task-specific simulator and did not handle load-dependent \ac{NFC} hydraulics or transfer across excavators.

Overall, model-based and data-driven motion controllers can be accurate, but most assume one hydraulic topology, abstract hydraulics as joint actuation or need machine-specific data, tuning or simulation.
They do not treat \ac{LS} and \ac{NFC} machines in a unified way.
Model-free \ac{RL} can be robust, but has not shown centimeter-level accuracy in deep heavy-duty grading.
None of these works provides a short retrofit calibration procedure for hydraulic adaptation across excavators.

\subsection{Grading and Surface Flattening}
Several works address grading and surface flattening with hydraulic excavators. 
\citet{Xu2014AutomatedGrading} proposed an automated grading controller based on Inverse Kinematics and PI hydraulic control.
Coupling the PI controllers can lead to improved precision as shown by \citet{Wang_Zheng_Yu_Zhou_Shao_2016}.
\citet{Lee_Bae_Hong_2013} formulated contour control for leveling tasks based on geometric path-following of the bucket edge by decoupling the target surface objective from pure position control.
More recent work uses contouring-error formulations and disturbance observers to achieve high-accuracy flattening: \citet{Dao_Na_Nguyen_Ahn_2021} presents contouring control for surface flattening and later extends the approach with active disturbance rejection and sliding-mode observation to better cope with external disturbances and model mismatch~\cite{Dao_Ahn_2022}. 
These techniques are evaluated in simulation only.
\citet{Yang_Zhang_Hong_Chen_Yang_Wang_Cao_2022} present a leveling controller for hydraulic excavators that combines a force pre-load term with cross-coupling compensation to synchronize cylinder motions on a small \ac{LS} excavator.

\begin{table*}[t]
    \centering
    \caption{Comparison to prior grading and hydraulic motion-control methods.}
    \label{tab:grading_comparison}
    \setlength{\tabcolsep}{4pt}
    \renewcommand{\arraystretch}{1.0}
    \footnotesize
    \begin{tabularx}{\textwidth}{@{}l l c c l c c@{}}
        \toprule
        \textbf{Ref.} & \textbf{Technology} & \textbf{Val.} & \textbf{Performance} & \textbf{Machine specificity} & \textbf{Arch.} \\
        \midrule
        \cite{Xu2014AutomatedGrading} 
            & IK + PI 
            & Sim. 
            & $\sim$4\,cm in sim, slow
            & single-machine model 
            & simplified 
            \\
            
        \cite{Dao_Na_Nguyen_Ahn_2021} 
            & Contouring + ESO 
            & Sim. 
            & N/A 
            & simulation-only
            & not considered
            \\
            
        \cite{Dao_Ahn_2022} 
            & ADRC + SMO contouring 
            & Sim. 
            & N/A
            & simulation-only
            & not considered 
            \\

        \cite{Kim_Kim_2025} 
            & Hybrid model-based + RL 
            & Sim. 
            & \SI{2.3}{\cm} 
            & tuned one simulated 30\,t excavator
            & in-air LS
            \\

\midrule
            
        \cite{Yang_Zhang_Hong_Chen_Yang_Wang_Cao_2022} 
            & Force preload + cross-coupling 
            & Real 
            & light grad. only: \SI{1}{\cm}
            & tuned on \SI{6}{\tonne} \ac{LS} excavator 
            & LS
            \\

        \cite{Wang_Zheng_Yu_Zhou_Shao_2016} 
            & Coupled non-linear PI 
            & Real 
            & in-air only: $\sim$2--9\,cm 
            & tuned on one machine
            & N/A
            \\
            
        \cite{Lee_Bae_Hong_2013} 
            & Contour following + PI
            & Real 
            & light grad. only: \SI{2.34}{\cm} avg
            & tuned on \SI{1.5}{\tonne} machine
            & N/A
            \\
            
        \cite{Lee_Choi_Kim_Moon_Kim_Lee_2022} 
            & Learned model inversion 
            & Real 
            & $<\!2$\,cm path RMSE 
            & several hours of machine-specific data
            & unclear
            \\
        \cite{Msaad_Cecchin_Demir_Fagiano_2025} 
            & Data-driven nonlinear MPC
            & Real 
            & cyl vel tracking only
            & identified on one excavator
            & LS
            \\
            
        \cite{trimbleGradeControl,catGrade,leicaMC1} 
            & Commercial, Proprietary
            & Real
            & $>$ \SI{4}{\cm} or light grading only
            & retrofittable
            & various 
            \\

        \midrule
        \textbf{Ours} 
            & \textbf{Hydraulic FF + PID + MPC} 
            & \textbf{Real} 
            & \textbf{1.4 / 1.8\,cm RMSE} 
            & \textbf{$\sim$20\,min, explicit calibration} 
            & \textbf{\ac{LS} + \ac{NFC}} 
            \\
        \bottomrule
    \end{tabularx}
    
    \vspace{1mm}
    \raggedright
    \footnotesize
    \textit{Val.}: validation type. 
    \textit{Arch.}: whether the method explicitly addresses hydraulic architectures. 
    N/A: not available in public description.
\end{table*}
 
Commercial solutions such as Trimble Earthworks and earlier GCS900 systems provide GNSS- and total-station-based guidance and semi-automatic boom, stick, and bucket control for excavator grading~\cite{trimbleGradeControl}. 
Caterpillar’s Cat Grade Control offers assistive modes to reach design surfaces and slopes and is tightly integrated with machine telematics and payload systems~\cite{catGrade}. 
Leica’s MC1 platform provides 2D machine control for multiple earthmoving machines, including excavators, and focuses on unified user interfaces and integration with survey equipment~\cite{leicaMC1}.
Details of the hydraulic control strategies and their adaptation to different hydraulic architectures are typically not disclosed in public documentation. 
However, as construction robotics is a more mature field, commercial solutions tend to match the performance of what is presented in literature.
Empirically, these systems perform well in light grading and finishing, but are limited in deeply loaded cuts and their use of the full hydraulic torque envelope.

All methods operate mainly in the kinematic or joint-velocity domain. 
They generally assume relatively light terrain interaction, do not explicitly model hydraulic dynamics or load-dependent effects, and ignore differences in joint response to input commands.
In contrast, our controller is retrofittable, hydraulics-aware, and machine-agnostic for heavy-duty high-precision grading.

Table~\ref{tab:grading_comparison} compares previous methods by validation type, performance, machine specificity, and hydraulic handling used.
Because many prior methods require vendor-level access, large datasets, extensive tuning, tuned simulation, or assumptions that do not hold on both excavators, we use a transferable commercial system deployed on the same machine as the practical baseline.
To the best of our knowledge, no prior method combines real in-soil validation, sub-\SI{2}{\cm} RMSE, short calibration, cross-machine deployment, and explicit handling of soft \ac{NFC} actuation.

\section{Controller Architecture}
\begin{figure}
    \centering
    \includegraphics[width=1.0\linewidth]{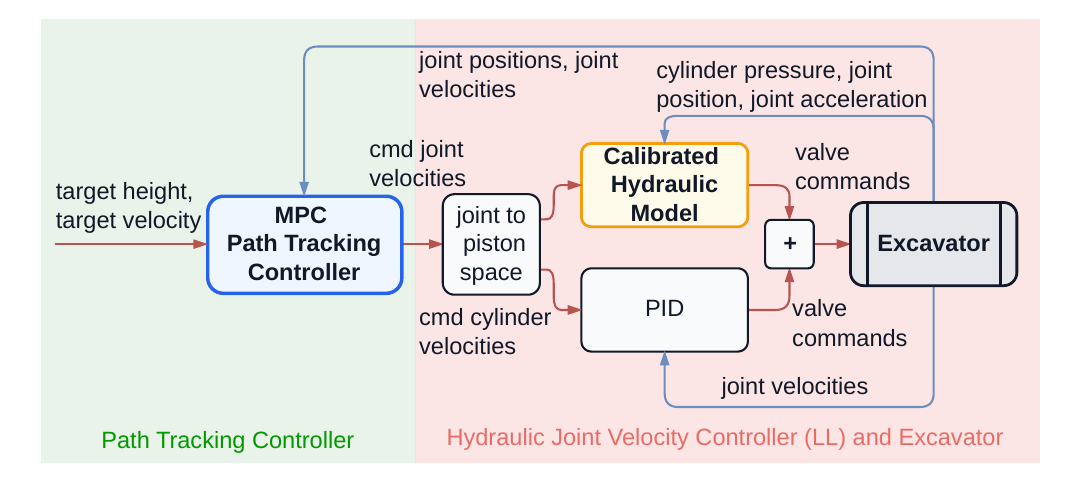}
    \vspace{-20pt}
    \caption{Flow graph of the grading controller and split between path tracking and joint velocity stage. The user sets the design surface and \ac{EE} speed. The MPC path-tracking controller outputs joint velocities, which are tracked by the hydraulic controller consisting of a hydraulic model and a parallel PID element.}
    \label{fig:flowgraph}
\end{figure}

The presented controller executes a single grading pass in the soil until the end of travel is reached or the machine stalls at maximum function pressure.
Our control pipeline receives as input from the user or a higher-level planner the height and inclination of the design surface, along with the desired grading velocity in \ac{EE} space.
To keep the motion controller adaptable across different machine types, we separate machine-agnostic path tracking from hydraulics-aware joint velocity control, as shown in Figure~\ref{fig:flowgraph}.

\subsection{Joint Velocity Controller}
Our core innovation includes a hydraulic-aware joint velocity controller that tracks target joint rates by computing a valve command or pilot-stage current.
Since joint-level effects occur at the cylinder level and are more linear there, we map the target joint rates [\SI{}{rad\per\s}] to cylinder speeds [\SI{}{\m\per\s}] via the closed-loop kinematics of each joint, as shown by \citet{Werner2025}. 
The target cylinder velocity is then fed to a calibrated hydraulic model as a \ac{FF} element and to a parallel PID controller.
During calibration, an appropriate hydraulic model is selected, enabling our method to generalize across hydraulic control loops such as \ac{LS} and \ac{NFC}.
Different hydraulic architectures vary significantly in their response to joint loading, making this adaptation vital.
The PID controller compensates for unmodeled effects and disturbances, such as flow sharing, pump boost, or recuperation, while most of the command comes from the calibrated, load-aware hydraulic \ac{FF} model.
While cross-coupling effects are present on all architectures, we found them to be compensatable with just the PID for our target speed and precision.
Implemented as a multidimensional \ac{LUT}, our hydraulic model maps desired cylinder velocities or function flows to valve commands using other measurements such as cylinder back pressure and joint acceleration. 
It acts as a \ac{MISO} element and is calibrated for each joint individually. 
Finally, the \ac{FF} and PID signals are combined and sent to the valves.
\subsubsection{Hydraulic Model Load Sensing}
Due to the load-aware and compensating behavior of \ac{LS} hydraulics, the calibrated hydraulic model can be designed in a straightforward way. 
A 1:1 \ac{LUT}, mapping from desired oil flow to valve command, calibrated on the machine as described in Section~\ref{sec:calib_ls} is sufficient.
Appendix~\ref{apx:lsHyd} explains the hydraulic mechanisms behind this model.

\subsubsection{Hydraulic Model \ac{NFC}}
\label{sec:hydModNFC}
\ac{NFC} hydraulics differ from \ac{LS} in how function loading affects motion~\cite{Li_Sun_Ding_Zhang_Yuan_Wu_2022}. 
Appendix~\ref{apx:nfcHyd} describes the more complex relationship between function load and flow in \ac{NFC} systems.
It also describes the underlying hydraulic model, that we calibrate as shown in Section~\ref{sec:calib_nfc} for the hydraulic-aware feed-forward element in our load-sensitive controller.
The simplified model is our key contribution, as it lets us calibrate and model NFC hydraulics with a task-specific goal. Hydraulic modeling is otherwise practically unviable for researchers due to the unavailability of core information from manufacturers and the difficulty of characterizing individual components, which requires full disassembly.

\subsection{Path Tracking Controller}
\label{sec:mpc}
Our path tracking controller uses the machine kinematics and transient behavior of the plant, which includes our joint velocity controller and the machine itself. 
A standard \ac{MPC} formulation is deployed, which uses a per-joint second-order transfer function with added dead-time delay as a system model.
By accounting for joint transients, the \ac{MPC} path tracking controller commands joint velocities that avoid the typical overshoot of the target surface at the start of a grading pass. 
We neglect joint coupling effects in favor of execution time, since they have shown limited influence on joint dynamics at the speeds targeted by this work. 
Future work could include them by explicitly modeling them in the \ac{MPC} formulation.

The \ac{MPC}'s objective is to execute a single grading pass along the design surface while respecting joint and actuator limits, outputting target joint rates for the joint velocity controller.
As an input, it gets the design surface height $h^\star$, inclination $\alpha^\star$ and \ac{EE} velocity $v_x^\star$.
The controller is formulated as a discrete-time nonlinear \ac{MPC} in joint space with a task-space tracking objective, where the \ac{EE} pose and twist are predicted through forward kinematics.

\paragraph{Prediction model.}
The \ac{MPC} state models the dominant closed-loop transient of each joint, including the calibrated hydraulic feedforward element, PID velocity loop and hydro-mechanical motion of the machine.
Each joint $i$ is approximated as a second-order system in velocity with gain $K_i$, damping ratio $\zeta_i$, natural frequency $\omega_{n,i}$, acceleration $a_i$ and dead time $\tau_i$ acting on the commanded joint velocity $u_i$ at time $t$:
\begin{align}
\dot{a}_i &= -2\zeta_i\omega_{n,i}\,a_i - \omega_{n,i}^2\,v_i + K_i\omega_{n,i}^2\,u_i(t-\tau_i),
\label{eq:mpc_joint_model}
\end{align}
with $v_i=\dot{\theta}_i$ and $a_i=\ddot{\theta}_i$. 
For the controlled joints, we define joint positions $\bm{\theta}\in\mathbb{R}^3$, velocities $\dot{\bm{\theta}}\in\mathbb{R}^3$, and accelerations $\ddot{\bm{\theta}}\in\mathbb{R}^3$.
The control input is the commanded joint velocity $\bm{u} = \dot{\bm{\theta}}_{\mathrm{cmd}} \in \mathbb{R}^3$.
The model is discretized at $\Delta t=\SI{0.1}{\second}$, matching the controller update rate.

Dead time is represented using a discrete input buffer of length $N_d$ that is part of the \ac{MPC} state.
Let $\bm{u}_{d,j}\in\mathbb{R}^3$ denote the $j$-th delayed-input state, with $j=1$ the most recent, and $(\cdot)^{+}$ denoting the next-step value.
The buffer dynamics are
\begin{align}
\bm{u}_{d,1}^{+} &= \bm{u}, \qquad\bm{u}_{d,j}^{+} = \bm{u}_{d,j-1}, \qquad j=2,\dots,N_d.
\label{eq:mpc_delay}
\end{align}
For joint $i$, the delayed input used in~\eqref{eq:mpc_joint_model} is selected as $u_i(t-\tau_i)\approx (\bm{u}_{d,d_i})_i$ with
\begin{align}
d_i = \mathrm{clip}\!\left(\left\lfloor \frac{\tau_i}{\Delta t}\right\rfloor,\,1,\,N_d\right),
\end{align}
where $\lfloor\cdot\rfloor$ is the floor operator and $\mathrm{clip}(\cdot,1,N_d)$ saturates the value to the interval $[1,N_d]$.

\paragraph{Task-space quantities.}
The \ac{MPC} cost is defined in task space. 
The excavator forward kinematics are evaluated inside the optimization using the predicted joint states. 
Specifically, the full configuration used for kinematics is
\begin{align}
\bm{q}_{\mathrm{full}} =
\begin{bmatrix}
\theta_{\mathrm{cab}}  & \theta_{\mathrm{boom}} & \theta_{\mathrm{stick}} & \theta_{\mathrm{bucket}}
\end{bmatrix}^\top ,
\end{align}
where $\theta_{\mathrm{cab}}$ is the measured cabin pitch.
For machines with a telescopic stick, stick extension measurements $\ell_{\mathrm{tele}}$ can be added to the state.
From $\bm{q}_{\mathrm{full}}$ and $\dot{\bm{q}}_{\mathrm{full}}$ we obtain the predicted \ac{EE} state $\bm{\pi}$ including  position $\bm{p}_{\ac{EE}}(\cdot)$, \ac{EE} linear velocity $\bm{v}_{\ac{EE}}(\cdot)$, and \ac{EE} angular velocity $\bm{\omega}_{\ac{EE}}(\cdot)$, all expressed in a local gravity-aligned frame.
The bucket pitch angle $\phi_{\ac{EE}}$ is extracted from the predicted \ac{EE} rotation matrix.

\paragraph{Optimization problem and cost function.}
At each control step, the \ac{MPC} solves a finite-horizon nonlinear program in receding-horizon fashion:
\begin{align}
\min_{\{\bm{u}_k\}_{k=0}^{N-1}} \quad
& \sum_{k=0}^{N-1} \ell(\bm{x}_k,\bm{u}_k,\bm{\pi}_k) + m(\bm{x}_N,\bm{\pi}_N) \\
\text{s.t.}\quad
& \bm{x}_{k+1} = f(\bm{x}_k,\bm{u}_k,\bm{\pi}_k), \\
& \bm{x}_0 = \hat{\bm{x}}(t), \qquad \bm{u}_k \in \mathcal{U}, \qquad \bm{\theta}_k \in \Theta .
\end{align}
Here, $\bm{x}_k$ is the \ac{MPC} state at step $k$ (containing the joint states and the delayed-input buffer), and $\hat{\bm{x}}(t)$ is the estimated current state at time $t$.
The mapping $f(\cdot)$ denotes the (discretized) state-transition function induced by~\eqref{eq:mpc_joint_model} and the delay-buffer dynamics~\eqref{eq:mpc_delay}.
The stage cost $\ell(\cdot)$ and terminal cost $m(\cdot)$ are scalar-valued functions.
The admissible input set $\mathcal{U}$ encodes joint-rate limits, and the admissible set $\Theta$ encodes joint-position limits.

For slope grading, the running cost penalizes deviation from the desired forward velocity, enforces near-zero vertical \ac{EE} velocity and pitch rate, regulates the \ac{EE} height to the target plane, and regulates the blade pitch:
\begin{align}
\ell(\bm{x}_k,\bm{u}_k,\bm{\pi}_k) &=
w_{vx}\left(v_{\ac{EE},x}(\bm{x}_k) - v_x^\star\right)^2\nonumber\\
&+ w_{vz}\left(v_{\ac{EE},z}(\bm{x}_k) - 0\right)^2\nonumber\\
&+ w_{\omega}\left(\omega_{\ac{EE},y}(\bm{x}_k) - 0\right)^2 \nonumber\\
&+ w_{\phi}\left(\phi_{\ac{EE}}(\bm{x}_k) - \phi^\star\right)^2\nonumber\\
&+ w_{h}\left(p_{\ac{EE},z}(\bm{x}_k) - (h^\star + \alpha^\star p_{\ac{EE},x}(\bm{x}_k)) \right)^2\nonumber\\
&+ w_{\Delta u}\left\lVert \bm{u}_k - \bm{u}_{k-1}\right\rVert_2^2\nonumber.
\end{align}
The increment penalty on $\bm{u}_k-\bm{u}_{k-1}$ promotes smooth joint velocity commands and reduces excitation of unmodeled effects in the low-level hydraulic loop. The terminal cost focuses on converging to the target height,
\begin{align}
m(\bm{x}_N,\bm{\pi}_N) = w_{h,N}\left(p_{\ac{EE},z}(\bm{x}_N) - \left(h^\star + \alpha^\star p_{\ac{EE},x}(\bm{x}_N)\right)\right)^2 .
\label{eq:mpc_terminal_cost}
\end{align}
In the implementation, $w_{\phi}$ is set to zero when the bucket joint is disabled.

\paragraph{Constraints.}
We enforce actuator and joint limits through the constraint sets $\mathcal{U}$ and $\Theta$.
\begin{align}
\bm{u}_{\min} \le \bm{u}_k \le \bm{u}_{\max}, \quad
\bm{\theta}_{\min} \le \bm{\theta}_k \le \bm{\theta}_{\max}.
\end{align}
When bucket motion is disabled, we tighten the bounds at each step to the current measured value and constrain the bucket velocity, acceleration, and input to zero.
This preserves a fixed state dimension while enforcing the desired reduced actuation.

\paragraph{Real-time implementation.}
The \ac{MPC} is solved at \SI{10}{\hertz} with horizon length $N = 20$. 
The solver receives the current joint state estimate $\hat{\bm{x}}(t)$ and the delayed input buffer from the previously applied commands. 
The commanded task-space velocity $\bm{v}^\star$ is treated as a time-varying parameter and updated at each step. 
In \ac{MPC} fashion, only the first control input is executed in each tick.

\section{Calibration}
This section describes the calibration of all free parameters in the controller pipeline, which is our second major contribution.
Our calibration procedure takes about \SI{20}{\min} and requires no disassembly of the system. 
Each joint is calibrated sequentially and in isolation, since coupling effects are not included in the model, starting with the joint velocity controller and finishing with the path tracking controller.
For the joint velocity controller, we calibrate our hydraulic \ac{FF} model as a \ac{LUT}.
Lookup values are directly sampled on \ac{LS} machines and derived from our variable displacement pump- and function orifice model on \ac{NFC} hydraulics according to Section~\ref{sec:hydModNFC} and detailed in~\ref{apx:nfcHyd}. After calibrating the hydraulic \ac{FF} model (Sections~\ref{sec:calib_ls} and~\ref{sec:calib_nfc}), the PID controller is tuned using standard methods~\cite{ziegler1942optimum}. 
For the path-tracking controller, we identify the transient response of the plant, containing the \ac{FF} element, PID controller, and hydraulic system.

In order to perform the calibration in the more linear cylinder space, we convert between cylinder and joint space using the position-dependent sensitivity factor $\boldsymbol{\gamma}(\boldsymbol{\Theta})$.
It maps cylinder forces $\mathbf{f}$ to joint torques $\boldsymbol{\tau}$ and cylinder velocities $\mathbf{v}$ to joint rates $\dot{\boldsymbol{\Theta}}$. 
The measured cylinder force $\mathbf{f_m}$ is computed from pressure readings on the $a$ and $b$ sides together with the corresponding plunger areas, as shown in Equation~\eqref{eq:pres2force} for one cylinder.
\begin{align}
    f_m &= p_a A_a - p_b A_b \label{eq:pres2force}
\end{align}

\subsection{Load Sensing}
\label{sec:calib_ls}
\begin{figure}
    \centering
    \includegraphics[width=0.8\linewidth]{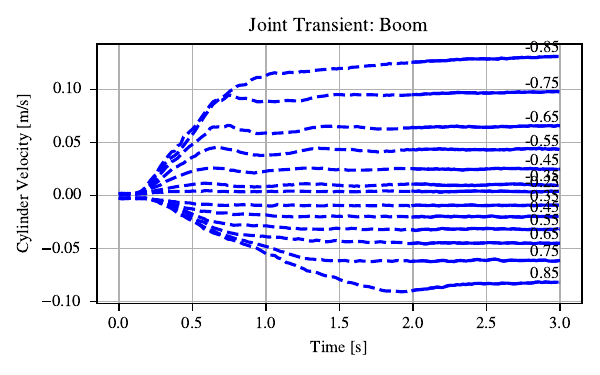}
    \caption{\ac{FF} data collection for the Boom joint. Normalized commands from -0.85 to 0.85 in steps of 0.1 with the resulting joint velocity response and steady-state part shown as a solid line. Data from Menzi Muck M445.}
    \label{fig:boomJointTransient}
\end{figure}

\begin{figure}
    \centering
    \includegraphics[width=0.8\linewidth]{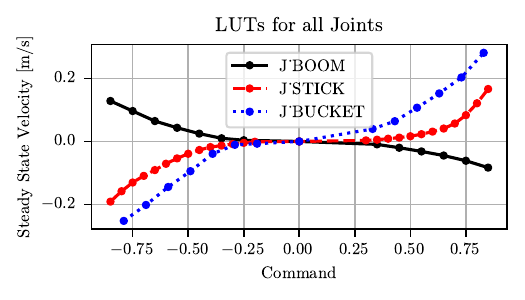}
    \caption{Hydraulic \ac{FF} model of the \ac{LS} M445. \ac{LS} enables a direct mapping between cylinder velocity and command.}
    \label{fig:m4luts}
\end{figure}

As \ac{LS} hydraulics show no significant load-dependent behavior, calibration reduces to the mapping from valve command $u$ to cylinder velocity $v$. 
We collect a dataset that spans the operational range of valve commands with small increments. 
Figure~\ref{fig:boomJointTransient} shows the measured cylinder responses for normalized commands in $[-0.85,\dots,0.85]$.
The average velocity over the third second after applying the command is used as the steady-state value.
We then apply linear interpolation between sampled velocities $v$ to obtain the final lookup function, as shown in Figure~\ref{fig:m4luts}.

\subsection{Negative Flow Control}
\label{sec:calib_nfc}
We propose a novel procedure for calibrating the \ac{FF} element in \ac{NFC} hydraulics, taking the load-dependent behavior into consideration.
Our method outputs a 2D \ac{LUT} that maps the desired cylinder speed and the current cylinder force to a valve command. 
Since it is not feasible to excite a fine grid of all speed and load combinations for direct calibration, we derive the \ac{LUT} from our simplified hydraulic model in Section~\ref{sec:hydModNFC}, \ref{apx:nfcHyd}, thus enabling straightforward calibration of \ac{NFC} hydraulics.
We identify the dependence of valve command on pump pressure and the free parameters of the function orifice in the \ac{DCV}.

\subsubsection{Inertia Compensation}
\begin{figure}
    \centering
    \includegraphics[width=1.0\linewidth]{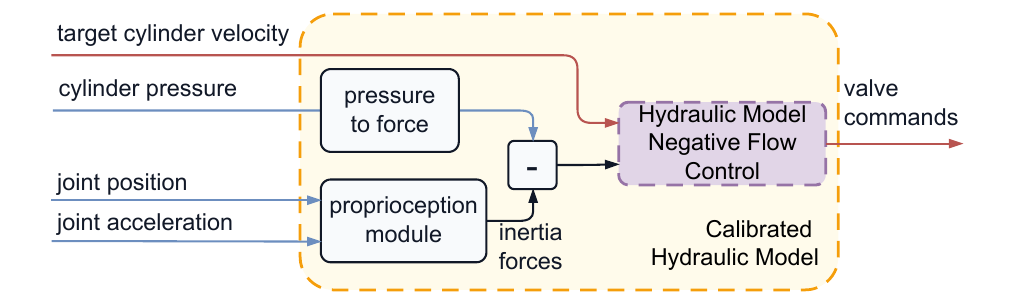}
    \vspace{-10pt}
    \caption{Flow chart of the \ac{FF} element for \ac{NFC} hydraulics. Inertia compensation for removal of system torques.}
    \label{fig:inertiaFlow}
\end{figure}

The pressure-derived cylinder force includes internal system forces and external forces from shovel–ground interaction. 
System forces depend on joint position (gravity), velocity (friction), and acceleration (inertia).
Gravity and friction create a quasi-static load, that is therefore compensated for by the \ac{FF} element. 
On the other hand, inertia-induced forces can create a positive feedback loop: a higher valve command increases acceleration, which causes a spike in reaction cylinder force, which then leads the \ac{FF} element to command an even higher valve command in order to compensate for the perceived increase in momentary load. 
This can make transients highly unstable.
We therefore estimate the inertia forces at the cylinder as described by \citet{Werner2025} and use inertia-compensated forces for the \ac{NFC} \ac{FF} element as shown in Figure~\ref{fig:inertiaFlow}.

\subsubsection{Hydraulic Model}
From the inertia-compensated cylinder forces, we compute an artificial cylinder back pressure at the function line as in Equation~\eqref{eq:pseudoBackP}, where $A_{\mathrm{articulation}}$ is the effective area of the active cylinder side in the calibrated direction.
Since external loading may act opposite to the articulation, the resulting function pseudo back pressure $P_f$ can also be negative.
\begin{align}
    P_{f} &= (f_m - f_{\mathrm{inertia}}) / A_{\mathrm{articulation}}
    \label{eq:pseudoBackP}
\end{align}

We model the pump's pressure loop setpoint with respect to the \ac{DCV} command $x$ as a \ac{LUT} with linear interpolation $P_p(x)$.
Furthermore, we calibrate the relationship between valve command $x$ and hydraulic resistance of the variable in-line orifice of the \ac{DCV}, represented by $R_3$ in~\ref{apx:nfcHyd}.
The resistance of the variable orifice in the \ac{DCV} is modeled by Equation~\eqref{eq:varOrifice},
\begin{align}
    R(x) &= \frac{a}{b \cdot (x + c)^2}
    \label{eq:varOrifice}
\end{align}
where the free parameters $a, b$ and $c$ are fitted from measurements of the volume flow into the function $Q_{\text{func}}$ and the function-side pseudo back pressure $P_f$.
The pressure over $R_3$ is then given by $P = P_p(x) - P_f$.
The orifice resistance $R$ is a lumped parameter derived from the turbulent orifice equation~\eqref{eq:orifice}, which is simplified to~\eqref{eq:orifice_simp}.
\begin{align}
    \Delta P &= \frac{Q^{2}\rho}{2C_{d}^{2}A^{2}}
    \label{eq:orifice}\\
    \Delta P &= R \cdot Q^2
    \label{eq:orifice_simp}
\end{align}
In practice, identification of $R$ requires a dataset with different joint loadings for the same valve command $x$.
\paragraph{Stick, Bucket}
For the \verb|Stick| and \verb|Bucket| joints we command a constant valve opening and progressively stall the bucket in the ground until standstill, which increases the joint load.
This motion generates trajectories of increasing $P_f$ and decreasing $Q_f$ that we use to fit the parameter $R_{\mathrm{test},x}$ for each test command $x$.

\paragraph{Boom}
Since for the lifting direction of the \verb|Boom| joint progressive stalling is not feasible, we record trajectories with different joint loads at the same command.
The configurations curled, erected, and erected with a full bucket are sufficient for calibration.
Each configuration yields one data point that we use in the same way as the trajectories of the previous joints.

\subsubsection{Lookup Table}
We fit Equation~\eqref{eq:varOrifice} to the experimentally determined $R_{\text{test},x}$ values with a logarithmic cost to preserve precision for small $R$ as shown in Equation~\eqref{eq:optim_R}.
\begin{align}
    \min_{a, b, c} \sum_{x \in x_{\text{test}}} \bigl[\left|\log(R(x)) - \log(R_{\text{test}, x})\right|\bigr]
    \label{eq:optim_R}
\end{align}

\begin{figure}
    \centering
    \includegraphics[width=0.8\linewidth]{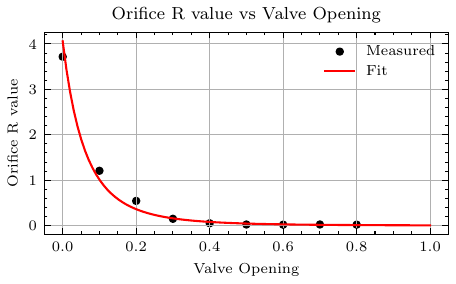}
    \vspace{-10pt}
    \caption{Identified orifice function for Stick. It maps hydraulic resistance to the valve command.}
    \label{fig:stickOrificeFunc}
\end{figure}

\begin{figure}
    \centering
    \includegraphics[width=0.8\linewidth]{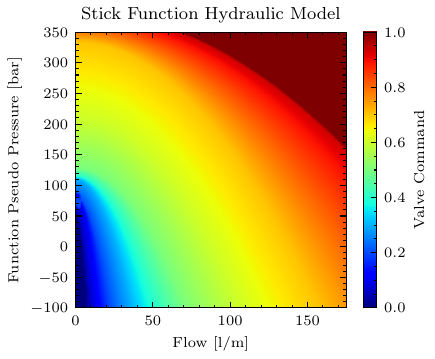}
    \vspace{-10pt}
    \caption{Hydraulic model for the Stick function on CASE250. Load-dependent mapping from valve command to flow (cylinder speed).}
    \label{fig:stickHydModel}
\end{figure}

The identified orifice model $R(x)$ shown in Figure~\ref{fig:stickOrificeFunc} and pump model $P_p(x)$ are inverted to form a two-dimensional \ac{LUT} that maps desired flow and current pseudo back pressure $P_f$ to a valve command $x$.
This table is the hydraulic model for a negative flow control function and is shown in Figure~\ref{fig:stickHydModel}.

\subsection{MPC Model}
\label{sec:calib_mpc}
\begin{figure}
    \centering
    \includegraphics[width=0.8\linewidth]{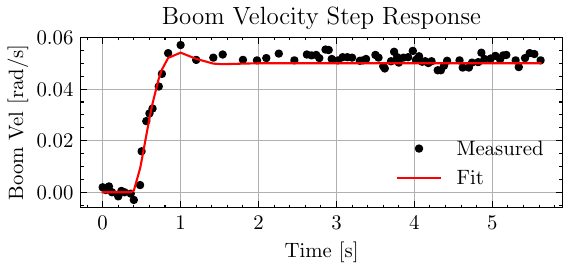}
    \vspace{-10pt}
    \caption{CASE250 Boom step response identification with a commanded joint rate of \SI{0.05}{\radian\per\sec}.}
    \label{fig:boomStepResp}
\end{figure}

The MPC controller is independent of the hydraulic model and assumes a stiff low-level controller, where small commands reach maximum function pressure at stall.
We calibrate the joint model from step responses with dead-time $\tau$, scalar $K$, damping $\zeta$ and natural frequency $\omega$ of the combined PID, hydraulic model and plant as shown in Figure~\ref{fig:boomStepResp}.

\section{Experiments}
\label{sec:experiments}
The presented method was tested and evaluated on a \SI{11.5}{\tonne} \ac{LS} excavator: Menzi Muck M445, and on a \SI{25}{\tonne} \ac{NFC} standard machine: CASE250.
The first experiments test hydraulic model accuracy under load.
The core quantitative evaluation is in-soil path tracking on CASE250 and M445 across cutting depths and external soil loads, including high-load passes near stall.
A separate CASE250 surface-finish test measures final terrain quality against a commercial system.
Preliminary analysis showed that the machine-specific commercial controller matches a straightforward PID+\ac{IK} controller for grading.

\subsection{Hydraulic FF}
\begin{figure}
    \centering
    \includegraphics[width=1.0\linewidth]{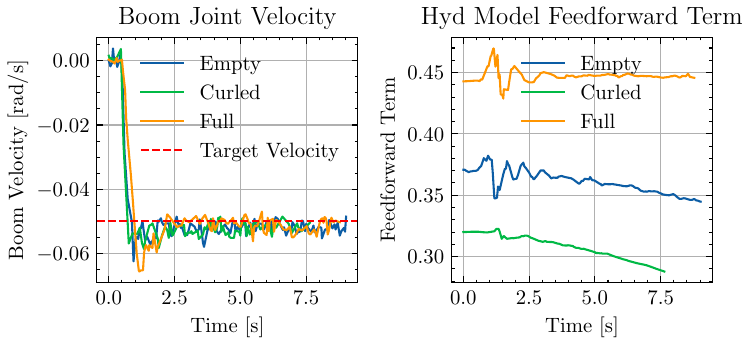}
    \vspace{-10pt}
    \caption{Contribution of calibrated NFC hydraulics \ac{FF} model. PID controller disabled. The pressure-dependent \ac{FF} term achieves the target joint velocity independent of the joint loading.}
    \label{fig:NFC_lifting}
\end{figure}

To validate the hydraulic \ac{FF} model, we benchmark the joint velocity tracking performance of the \ac{FF} element for different payloads and thus joint loadings.
On the \ac{LS} M445 this experiment only validates the calibrated \ac{LUT}, since the system reacts stiff to different loads.
A joint-velocity tracking accuracy of $\approx 2\%$ is achieved, confirming the \ac{LS} model and calibration.
On the CASE250 this experiment shows that the hydraulic model can compensate for the load-dependent joint dynamics.
Figure~\ref{fig:NFC_lifting} shows a \verb|Boom| lifting command of \SI{0.05}{\radian\per\s} for three payload scenarios.
The target joint velocity is accurately tracked by the load-dependent \ac{FF} term, including changing gravity loading during the lift, as visible in the \ac{FF} output.
Comparable performance is obtained over the full machine velocity range and payloads.

\subsection{Grading}
Our grading experiments test the full stack, including joint velocity control and \ac{MPC} path tracking.
We compare the precision of the proposed system with a commercially available semi-automatic grading controller and evaluate our proprioceptive path-tracking accuracy and soil-planing accuracy across multiple passes using a laser scanner.
All experiments, including those with the commercial system, use a blade velocity of approximately \SI{0.5}{\m\per\s} for comparability.
This velocity is close to that of an expert operator during fine grading, who removes less material but achieves similar surface accuracy.

\subsubsection{Path Tracking}
\begin{figure}
    \centering
    \includegraphics[width=0.8\linewidth]{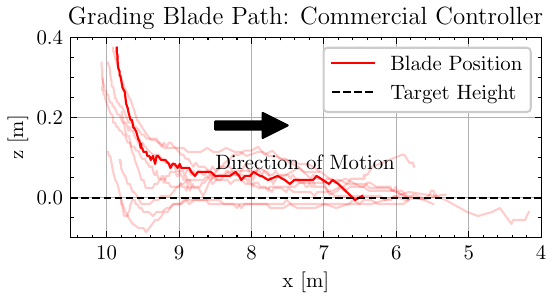}
    \vspace{-10pt}
    \caption{CASE250, Commercial Controller: Evaluation of 10 grading passes of varying depth in compacted soil. RMSE: \SI{4.7}{\cm} (\SI{9}{\m} and closer)}
    \label{fig:evalPathTrackLeica}
\end{figure}

\begin{figure}
    \centering
    \includegraphics[width=0.8\linewidth]{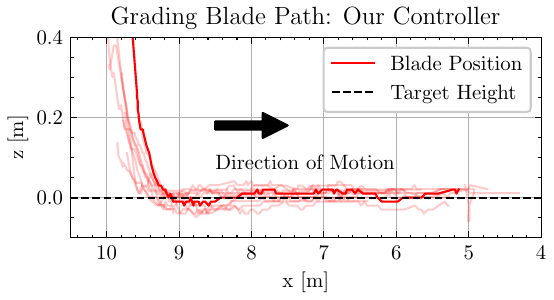}
    \vspace{-10pt}
    \caption{CASE250, Our Controller: Evaluation of 10 grading passes of varying depth in compacted soil. RMSE: \SI{1.8}{\cm} (\SI{9}{\m} and closer)}
    \label{fig:evalPathTrackOurs}
\end{figure}

\begin{figure}
    \centering
    \includegraphics[width=0.8\linewidth]{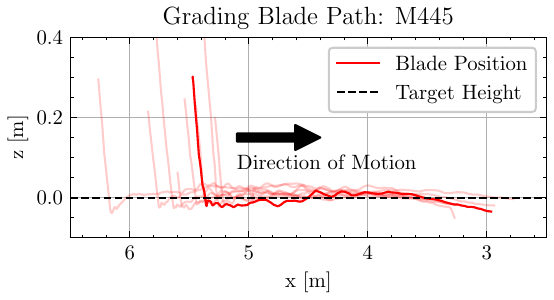}
    \vspace{-10pt}
    \caption{M445, Our Controller: Evaluation of 10 grading passes of varying depth in compacted soil. RMSE: \SI{1.4}{\cm} (\SI{5.2}{\m} and closer). Varying starting distances are a result of different extensions of the telescopic stick on M445.}
    \label{fig:evalPathTrackM4}
\end{figure}

We collect trajectories from light surface grading to deeper cuts, high-load passes, and near-stall cuts in compacted material from \SI{2}{\cm}  to \SI{40}{\cm} deep.
The deep paths in Figures~\ref{fig:evalPathTrackLeica}--\ref{fig:evalPathTrackM4} fill the bucket in less than one stroke.
We also test whether the machine reaches maximum torque and whether the blade is pushed away from the target surface.

\paragraph{Commercial (CASE250)}
Figure~\ref{fig:evalPathTrackLeica} shows the trajectory tracking of the commercial controller on the CASE250.
After the first meter of approach to the target, it tracks with an \textbf{RMSE of 4.7~cm}  and a maximum \textbf{path deviation of 17~cm} (\SI{9}{\m} and closer).
In hard soil the controller does not reach the target depth, while in light soil a substantial initial overshoot is visible.
Many trajectories stall before the end of travel at about \SI{5}{\m} with stick pressures far below the maximum function pressure of \SI{350}{\bar}.
In some hard soil conditions the shovel is pushed up at the end of travel, reducing accuracy close to the machine.

\paragraph{Ours (CASE250)}
As shown in Figure~\ref{fig:evalPathTrackOurs}, the presented controller achieves an \textbf{RMSE of 1.8~cm} and a maximum \textbf{path deviation of 5~cm} on the same machine and conditions.
It maintains high accuracy regardless of soil conditions and reliably reaches the target height after about one meter of approach.
Light grading passes show no initial overshoot, and in most cases the end of travel is reached.
When the machine stalls earlier, the stick pressure reaches the maximum function pressure of the machine.

\paragraph{Ours (M445)}
Figure~\ref{fig:evalPathTrackM4} shows our controller on the \ac{LS} Menzi Muck M445 with an \textbf{RMSE of 1.4~cm}\footnote{Much lower in-air. Precise enough to open a bottle. Check our supplementary \href{https://youtu.be/bCOMYbRWv5I}{video} for a visual demonstration of the achieved accuracy.} and a maximum \textbf{path deviation of 4~cm}.
On M445 the trajectories start at different distances due to the variable reach from the telescopic joint which is initialized at different lengths but kept fixed during episodes.
No loss of precision from varying cutting depths is visible, consistent with the CASE250 results.
The trajectory stalls at maximum function pressure and shows no significant initial overshoot.

\subsubsection{Surface Quality}
\begin{figure}
    \centering
    \includegraphics[width=0.9\linewidth]{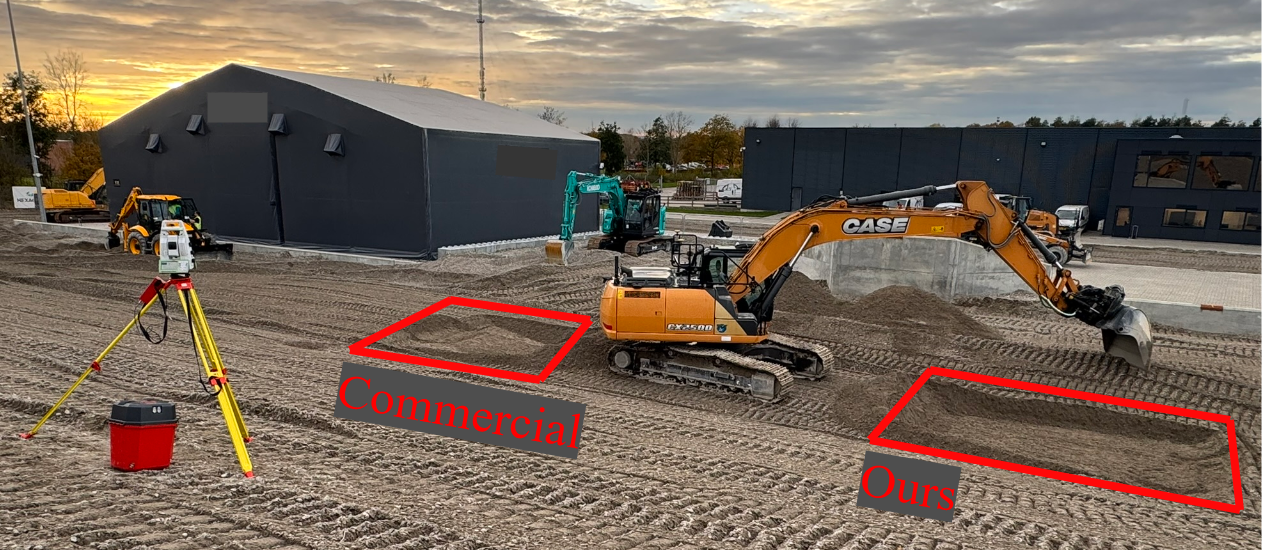}
    \caption{Experiment Setup for Surface Quality Measurements on CASE250. Fixed Leica MS60 scan used for evaluation of the surface quality.}
    \label{fig:exp:surfaceQual}
\end{figure}

\begin{figure}
    \centering
    \includegraphics[width=0.8\linewidth]{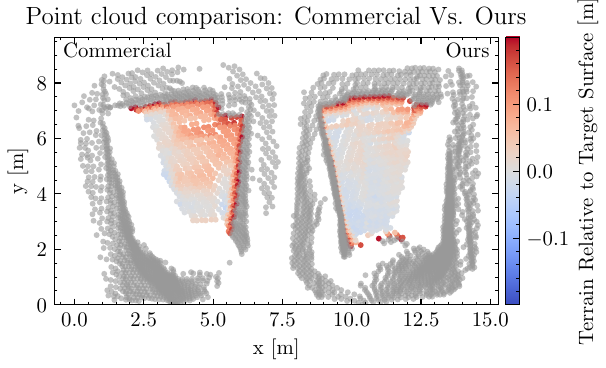}
    \vspace{-10pt}
    \caption{CASE250: Comparison of the two graded surfaces: target surface was not reached with the commercial system, higher precision and consistency from our method. Gray points are outside of the testing area and included for context, white areas are shadowed in the scan.}
    \label{fig:pitScan}
\end{figure}

This separate CASE250 test evaluates final surface finish, not path tracking, as seen in Figure~\ref{fig:exp:surfaceQual}.
The test is performed on compacted sand with a cutting depth of about \SI{25}{\cm}.
The target area is four shovel widths wide and graded from left to right, one pass per lateral position with about 20\% overlap (five passes total) to avoid spillage.
Surface quality is measured by laser scan with a Leica MS60 Multistation.
Figure~\ref{fig:pitScan} shows the two graded areas side by side, with color indicating deviation from the target surface and gray points outside the graded area.
The commercial controller does not reach the target height until the midpoint of the pass and shows large deviations and surface oscillations with magnitudes larger than \SI{10}{\cm}.
The presented controller reaches the target surface earlier, creates a flatter surface at the target height and does not oscillate.

\subsection{Discussion}
Our calibrated hydraulic \ac{FF} terms track joint velocity commands across payloads on \ac{LS} and \ac{NFC} machines.
Grading experiments show that the proposed controller reaches the target surface after about one meter and maintains low deviation in compacted material up to \SI{40}{\cm} cutting depth.
Our setup \textbf{outperforms the commercial controller by a factor of 2.6} (RMSE \SI{1.8}{\cm} vs.\ \SI{4.7}{\cm}) and avoids premature stalling by using maximum function pressure.
Laser scans confirm a flatter surface with fewer oscillations and earlier convergence to the target height.

Our once-per-machine retrofitting process of the presented controller can adapt to \ac{LS} and \ac{NFC} hydraulic machines.
It starts with the identification and tuning of the hydraulic-specific low-level controller, which includes inertia compensation for \ac{NFC} machines.
The transfer functions of the plant including machine and low-level controller are then identified and updated in the \ac{MPC} \ac{EE} space controller.
The whole process takes about \SI{20}{\min}.

\section{Future Work, Limitations, and Conclusion}
\subsection{Future Work and Limitations}
Future work will address tracking arbitrary shapes via a more expressive cost function and will include end-effector forces in Cartesian space in the \ac{MPC} objective to enable impedance-style control for proprioceptive compaction and redistribution of piles.
Limitations of this work are the applicability of our method to Positive Flow Control machines, which has not been evaluated, as no machine with this architecture was available to the authors.
The hydraulic model can be extended to capture coupling effects between functions, enabling faster motions at higher precision.

\subsection{Conclusion}
\label{sec:conclusion}
This work presents a retrofittable, machine-agnostic control stack for high-precision, heavy-duty grading with hydraulic excavators.
Our joint velocity controller targets multiple hydraulic architectures, including \ac{LS}, and for \ac{NFC} machines we formulate a novel load-aware model.
Furthermore, we formulate a practical calibration procedure that targets field deployment: it relies only on kinematic parameters and onboard sensing and runs in about \SI{20}{\min}. Experiments on a \SI{11.5}{\tonne} \ac{LS} and a \SI{25}{\tonne} \ac{NFC} excavator validate the \ac{FF} joint velocity accuracy and in-soil grading performance of the combined system.
Our system executes grading passes in compacted material with cutting depths up to \SI{40}{\cm}, respects actuator limits, and preserves smooth joint commands through explicit transient and dead-time modeling.
In heavy-duty grading trials, the proposed controller outperforms a state-of-the-art commercially available semi-automatic system on the same \ac{NFC} excavator and uses the full torque potential.
Overall, the proposed retrofittable controller achieves centimeter-level grading at 2.6 times better accuracy in compacted material and generalizes across common hydraulic architectures. 
Adoption of our grading framework can reduce the number of autonomous passes required in earthworks, resulting in substantial cost reductions.

\section*{Acknowledgments}
This project received funding from Hexagon AB and was supported by the Swiss National Science Foundation through the National Centre of Competence in Digital Fabrication (NCCR dfab).

\bibliographystyle{plainnat}
\bibliography{sn-bibliography}

@article{Aoshima_Servin_2024, title={\href{https://doi.org/10.1007/s11044-024-10005-5}{Examining the simulation-to-reality gap of a wheel loader digging in deformable terrain}}, ISSN={1573-272X}, DOI={10.1007/s11044-024-10005-5}, abstractNote={We investigate how well a physics-based simulator can replicate a real wheel loader performing bucket filling in a pile of soil. The comparison is made using field-test time series of the vehicle motion and actuation forces, loaded mass, and total work. The vehicle was modeled as a rigid multibody system with frictional contacts, driveline, and linear actuators. For the soil, we tested discrete-element models of different resolutions, with and without multiscale acceleration. The spatiotemporal resolution ranged between 50–400 mm and 2–500 ms, and the computational speed was between 1/10,000 to 5 times faster than real time. The simulation-to-reality gap was found to be around 10% and exhibited a weak dependence on the level of fidelity, e.g., compatible with real-time simulation. Furthermore, the sensitivity of an optimized force-feedback controller under transfer between different simulation domains was investigated. The domain bias was observed to cause a performance reduction of 5% despite the domain gap being about 15%.}, journal={Multibody System Dynamics}, author={Aoshima, Koji and Servin, Martin}, year={2024}, month=jul, language={en} }

@article{Egli_Hutter_2022, title={\href{https://doi.org/10.1109/LRA.2022.3152865}{A General Approach for the Automation of Hydraulic Excavator Arms Using Reinforcement Learning}}, volume={7}, ISSN={2377-3766}, DOI={10.1109/LRA.2022.3152865}, abstractNote={This article presents a general approach to derive an end effector trajectory tracking controller for highly nonlinear hydraulic excavator arms. Rather than requiring an analytical model of the system, we use a neural network model that is trained based on measurements collected during operation of the machine. The data-driven model effectively represents the actuator dynamics including the cylinder-to-joint-space conversion. Requiring only the distances between the individual joints, a simulation is set up to train a control policy using reinforcement learning (RL). The policy outputs pilot stage control commands that can be directly applied to the machine without further fine-tuning. The proposed approach is implemented on a Menzi Muck M545, a 12 mathrmt hydraulic excavator, and tested in different task space trajectory tracking scenarios, with and without soil interaction. Compared to a commercial grading controller, which requires laborious hand-tuning by expert engineers, the learned controller shows higher tracking accuracy, indicating that the achieved performance is sufficient for the practical application on construction sites and that the proposed approach opens a new avenue for future machine automation.}, number={2}, journal={IEEE Robotics and Automation Letters}, author={Egli, Pascal and Hutter, Marco}, year={2022}, month=apr, pages={5679–5686} }

@article{Jud19AutonomousFreeForm,
  title = {\href{https://doi.org/10.1109/LRA.2019.2925758}{Autonomous Free-Form Trenching Using a Walking Excavator}},
  author = {Jud, Dominic and Leemann, Philipp and Kerscher, Simon and Hutter, Marco},
  year = {2019},
  journal = {IEEE Robotics and Automation Letters},
  volume = {4},
  number = {4},
  pages = {3208--3215},
  urldate = {2024-02-25},
  langid = {english},
  doi={10.1109/LRA.2019.2925758}
}

@article{Lee_Choi_Kim_Moon_Kim_Lee_2022, title={\href{https://doi.org/10.1109/LRA.2022.3142389}{Precision Motion Control of Robotized Industrial Hydraulic Excavators via Data-Driven Model Inversion}}, volume={7}, ISSN={2377-3766}, DOI={10.1109/LRA.2022.3142389}, abstractNote={This work proposes a novel precision motion control framework of robotized industrial hydraulic excavators via data-driven model inversion. Rather than employing a single neural network to approximate the whole excavator dynamics, including input delays and dead-zones, we construct a physics-inspired data-driven model with a modular structure. The data-driven model is then inverted in a modular fashion which benefits the training speed. The data-driven model and its inversion are trained offline in a supervised manner using the real operational data since online learning methods can damage the machine and surroundings. The entire motion control framework consists of the data-driven model inversion that compensates for the excavator dynamics and the proportional control that determines the input of the model inversion to enhance the robustness. The framework is experimentally validated with a commercial 38-ton class hydraulic excavator for digging and grading tasks, achieving a precise control performance (i.e., root-mean-square of the path following error under 2 ;[rm cm]) even under severe soil interactions.}, number={2}, journal={IEEE Robotics and Automation Letters}, author={Lee, Minhyeong and Choi, Hyelim and Kim, ChangU and Moon, Jihyun and Kim, Dongmok and Lee, Dongjun}, year={2022}, month=apr, pages={1912–1919} }

@article{Spinelli_Egli_Nubert_Nan_Bleumer_Goegler_Brockes_Hofmann_Hutter_2024, title={\href{https://doi.org/10.1109/IROS58592.2024.10802199}{Reinforcement Learning Control for Autonomous Hydraulic Material Handling Machines with Underactuated Tools}}, ISSN={2153-0866}, DOI={10.1109/IROS58592.2024.10802199}, abstractNote={The precise and safe control of heavy material handling machines presents numerous challenges due to the hard-to-model hydraulically actuated joints and the need for collision-free trajectory planning with a free-swinging end-effector tool. In this work, we propose an RL-based controller that commands the cabin joint and the arm simultaneously. It is trained in a simulation combining data-driven modeling techniques with first-principles modeling. On the one hand, we employ a neural network model to capture the highly nonlinear dynamics of the upper carriage turn hydraulic motor, incorporating explicit pressure prediction to handle delays better. On the other hand, we model the arm as velocity-controllable and the free-swinging end-effector tool as a damped pendulum using first principles. This combined model enhances our simulation environment, enabling the training of RL controllers that can be directly transferred to the real machine. Designed to reach steady-state Cartesian targets, the RL controller learns to leverage the hydraulic dynamics to improve accuracy, maintain high speeds, and minimize end-effector tool oscillations. Our controller, tested on a mid-size prototype material handler, is more accurate than an inexperienced operator and causes fewer tool oscillations. It demonstrates competitive performance even compared to an experienced professional driver.}, journal={International Conference on Intelligent Robots and Systems}, author={Spinelli, Filippo A. and Egli, Pascal and Nubert, Julian and Nan, Fang and Bleumer, Thilo and Goegler, Patrick and Brockes, Stephan and Hofmann, Ferdinand and Hutter, Marco}, year={2024}, month=oct, pages={12694–12701} }

@article{Terenzi_Hutter_2024, title={\href{https://doi.org/10.1109/TFR.2024.3485037}{Toward Autonomous Excavation Planning}}, volume={1}, ISSN={2997-1101}, DOI={10.1109/TFR.2024.3485037}, abstractNote={Excavation plans are essential in construction projects, dictating the dirt disposal strategy and excavation sequence based on the final geometry and machinery available. While most construction processes rely heavily on coarse sequence planning and local execution planning driven by human expertise and intuition, fully automated planning tools are notably absent from the industry. This article introduces a fully autonomous excavation planning system. Initially, the site is mapped, followed by user selection of the desired excavation geometry. The system then invokes a global planner to determine the sequence of poses for the excavator, ensuring complete site coverage. For each pose, a local excavation planner decides how to move the soil around the machine, and a digging planner subsequently dictates the sequence of digging trajectories to complete a patch. We showcased our system by autonomously excavating the largest pit documented so far, achieving an average digging cycle time of roughly 30 s.}, journal={IEEE Transactions on Field Robotics}, author={Terenzi, Lorenzo and Hutter, Marco}, year={2024}, pages={292–317} }

@ARTICLE{10616135,
  author={Egli, Pascal and Terenzi, Lorenzo and Hutter, Marco},
  journal={IEEE Transactions on Field Robotics}, 
  title={\href{https://www.doi.org/10.1109/TFR.2024.3432508}{Reinforcement Learning-Based Bucket Filling for Autonomous Excavation}}, 
  year={2024},
  volume={1},
  number={},
  pages={170-191},
  doi={10.1109/TFR.2024.3432508}}

@misc{trimbleGradeControl,
  author = {Trimble Inc.},
  title = {Trimble Earthworks},
  howpublished = {\url{https://heavyindustry.trimble.com/en/products/grade-control-excavators}},
  note = {Accessed: 2025-11-10}
}

@misc{catGrade,
  author = {Caterpillar},
  title = {Cat Grade for Excavators},
  howpublished = {\url{https://www.cat.com/en_US/products/new/technology/grade/grade/15969804.html}},
  note = {Accessed: 2025-11-10}
}

@misc{leicaMC1,
  author = {Leica Geosystems},
  title = {Leica MC1},
  howpublished = {\url{https://leica-geosystems.com/en-us/products/machine-control-systems/software/leica-mc1}},
  note = {Accessed: 2025-11-10}
}

@article{Werner2025,
  title = {\href{https://doi.org/10.1007/s41693-025-00169-7}{Calibrated dynamic modeling for force and payload estimation in hydraulic machinery}},
  volume = {9},
  ISSN = {2509-8780},
  DOI = {10.1007/s41693-025-00169-7},
  number = {2},
  journal = {Construction Robotics},
  publisher = {Springer Science and Business Media LLC},
  author = {Werner,  Lennart and Eyschen,  Pol and Costello,  Sean and Micarelli,  Pierluigi and Hutter,  Marco},
  year = {2025},
  month = oct 
}

@article{Wang_Wang_2014, title={\href{https://doi.org/10.1016/j.autcon.2014.07.012}{Efficiency analysis and evaluation of energy-saving pressure-compensated circuit for hybrid hydraulic excavator}}, volume={47}, ISSN={0926-5805}, DOI={10.1016/j.autcon.2014.07.012}, abstractNote={A hydraulic cylinder driven scheme combining a pressure compensator and an energy recovery device together has been proposed to achieve good control operation and energy-saving capability simultaneously. In this paper, its efficiency characteristics are further investigated in order to provide analytical and experimental references to practical applications. Since an excavator owns multiple actuators, a general schematic configuration including hydraulic cylinders with and without energy recovery is developed and analyzed. Based on the analysis of energy losses in every conversion, component selections and possible improvements are discussed, and then the design flowchart and criteria of key parameters are also presented. Finally, experiments under different load and velocity conditions are implemented on a test bench. Energy distributions, recovery efficiencies and component efficiencies are all evaluated.}, journal={Automation in Construction}, author={Wang, Tao and Wang, Qingfeng}, year={2014}, month=nov, pages={62–68} }

@article{Li_Sun_Ding_Zhang_Yuan_Wu_2022, title={\href{https:doi.org/10.3390/pr10122482}{Review of Flow-Matching Technology for Hydraulic Systems}}, volume={10}, rights={http://creativecommons.org/licenses/by/3.0/}, ISSN={2227-9717}, DOI={10.3390/pr10122482}, abstractNote={The flow-matching problem of hydraulic systems is an important factor affecting the working performance and energy saving of hydraulic systems. According to the different flow-matching mechanisms, the flow-matching technology of hydraulic systems can be divided into three categories: positive flow-control technology, negative flow-control technology, and load-sensitive control technology. In this paper, the working mechanism of flow-matching technology and the cause of energy loss are analyzed, and the research results of flow matching are introduced from two aspects of energy saving and consumption reduction and system performance improvement. In the direction of energy saving and consumption reduction, the purposes of energy saving and consumption reduction are achieved by means of multi-way valve commutation, independent inlet and outlet control, parallel replacement of shuttle valve by a cylinder piston rod controlled by pilot pressure, change of hydraulic resistance of a pressure compensating valve, improvement of the power regulation range of a hydraulic pump, and potential energy recovery. In the direction of system performance, by means of flow-forecasting system pressure change, applying flow unsaturation real-time control idea, and combining electronic control technology with load-sensitive technology, the pressure drop during transmission process and the transmission signal lag are reduced, the speed regulation interval is enlarged, fine-tuning characteristics are improved, and the response speed is increased. The research results indicate that improving the structure and the control strategy of hydraulic systems and improving the flow-matching degree of a system to achieve global matching will be a future development trend.}, number={12}, journal={Processes}, publisher={Multidisciplinary Digital Publishing Institute}, author={Li, Ruichuan and Sun, Qiyou and Ding, Xinkai and Zhang, Yisheng and Yuan, Wentao and Wu, Tong}, year={2022}, month=dec, pages={2482}, language={en} }

@misc{insaneHydraulics,
  author = {Sergiy Sydorenko},
  title = {Negative Flow Control vs Load Sensing - Which One is Better?},
  howpublished = {\url{https://www.insanehydraulics.com/letstalk/nfcvsls.html}},
  note = {Accessed: 2025-12-04},
  year			= "2025"
}

@book{hydbook,
  editor		= "Timothy W. Dell",
  title			= "\href{https://www.g-w.com/hydraulic-systems-mobile-equipment-2023}{Hydraulic Systems for Mobile Equipment}",
  publisher		= "G-W Publisher",
  year			= "2023"
}

@article{ziegler1942optimum,
  title={Optimum settings for automatic controllers},
  author={Ziegler, John G and Nichols, Nathaniel B},
  journal={Transactions of the American society of mechanical engineers},
  volume={64},
  number={8},
  pages={759--765},
  year={1942},
  publisher={American Society of Mechanical Engineers}
}

@ARTICLE{10697977,
  author={Jang, Inkyu and Kim, Junha and Lee, Dongjae and Kim, Changhyeon and Oh, Changsuk and Kim, Youngbum and Woo, Sangwook and Sung, Heejee and Kim, H. Jin},
  journal={IEEE Robotics \& Automation Magazine}, 
  title={\href{https://www.doi.org/10.1109/MRA.2024.3400772}{Towards Fully Integrated Autonomous Excavation: Autonomous Excavator for Precise Earth Cutting and Onboard Landscape Inspection}}, 
  year={2025},
  volume={32},
  number={3},
  pages={88-102},
  doi={10.1109/MRA.2024.3400772}}

@article{Azar_Kamat, title={\href{https://www.itcon.org/2017/13}{Earthmoving equipment automation: a review of technical advances and future outlook}}, abstractNote={In the construction industry, the earthmoving sector is among the pioneers in adopting new sensing and information technologies to reduce operation costs, improve productivity, and enhance automation and safety. Fleet tracking and management systems, automated machine guidance and control, and proximity detection devices for accident warning are some examples of emerging products for earthmoving equipment. In addition to the commercial solutions, the research community actively develops and evaluates new systems in this area. This paper aims to critically review the related advances in this field. A three-phase literature review was carried out to investigate the innovations in industrial and academic research communities. Advances in six major industrial companies and a total of 102 related academic papers have been reviewed and discussed. Based on the application area and the function, current research works are divided into four categories: equipment tracking and fleet management, safety management, equipment pose estimation and machine control technology, and remote control and autonomous operation. The underlying technologies and methods used in these systems are discussed in detail. Finally, future research opportunities, based on the identified shortcomings and gaps in knowledge, are highlighted. In particular, the remote control and autonomous operation of earthmoving equipment are identified as the most underdeveloped and complicated areas, and the missing modules and research directions in these fields are discussed.}, author={Azar, Ehsan Rezazadeh and Kamat, Vineet R}, language={en}, url={https://www.itcon.org/2017/13}, volume={22}, pages={247-265},  journal={Tcon}, }

@article{Nguyen_Ha_2023, title={\href{https://doi.org/10.1017/S0263574722000339}{Robotic autonomous systems for earthmoving equipment operating in volatile conditions and teaming capacity: a survey}}, volume={41}, ISSN={0263-5747, 1469-8668}, DOI={10.1017/S0263574722000339}, abstractNote={There has been an increasing interest in the application of robotic autonomous systems (RASs) for construction and mining, particularly the use of RAS technologies to respond to the emergent issues for earthmoving equipment operating in volatile environments and for the need of multiplatform cooperation. Researchers and practitioners are in need of techniques and developments to deal with these challenges. To address this topic for earthmoving automation, this paper presents a comprehensive survey of significant contributions and recent advances, as reported in the literature, databases of professional societies, and technical documentation from the Original Equipment Manufacturers (OEM). In dealing with volatile environments, advances in sensing, communication and software, data analytics, as well as self-driving technologies can be made to work reliably and have drastically increased safety. It is envisaged that an automated earthmoving site within this decade will manifest the collaboration of bulldozers, graders, and excavators to undertake ground-based tasks without operators behind the cabin controls; in some cases, the machines will be without cabins. It is worth for relevant small- and medium-sized enterprises developing their products to meet the market demands in this area. The study also discusses on future directions for research and development to provide green solutions to earthmoving.}, number={2}, journal={Robotica}, author={Nguyen, Huynh A. D. and Ha, Quang P.}, year={2023}, month=feb, pages={486–510}, language={en} }

@inproceedings{Xu2014AutomatedGrading,
  author    = {Jiaqi Xu and Bradley Thompson and Hwan-Sik Yoon},
  title     = {\href{https://doi.org/10.4271/2014-01-2405}{Automated Grading Operation for Hydraulic Excavators}},
  booktitle = {SAE Technical Paper 2014-01-2405},
  year      = {2014},
  doi       = {10.4271/2014-01-2405}
}

@article{Lee_Bae_Hong_2013, title={\href{https://doi.org/10.1007/s12541-013-0278-5}{Contour control for leveling work with robotic excavator}}, volume={14}, ISSN={2005-4602}, DOI={10.1007/s12541-013-0278-5}, abstractNote={This paper presents electro-hydraulic servo systems of a robotic excavator with a contour control algorithm. It is very important to precisely move the bucket tip of the excavator to a desired trajectory. There have been many studies to accurately control the bucket dealing with the non-linearity in the hydraulic boom, arm, and bucket cylinders. Beginning with these conventional methods, a new method that focuses on keeping contours rather than just following position commands is presented. In the leveling work, for example, it is more important to maintain linear contour than chasing a position goal. The contour control shares the control effort to make the bucket stay on the path, while slightly sacrificing the position tracking accuracy. After the kinematics of the excavator system were analyzed, the contour control algorithm was developed. The algorithm was applied to leveling work with a real excavator. The experiments showed better performance than using position control alone.}, number={12}, journal={International Journal of Precision Engineering and Manufacturing}, author={Lee, Chang Seop and Bae, Jangho and Hong, Daehie}, year={2013}, month=dec, pages={2055–2060}, language={en},doi= {10.1007/s12541-013-0278-5} }

@article{Wang_Zheng_Yu_Zhou_Shao_2016, title={\href{https://doi.org/10.1016/j.autcon.2015.12.024}{Robotic excavator motion control using a nonlinear proportional-integral controller and cross-coupled pre-compensation}}, volume={64}, ISSN={09265805}, DOI={10.1016/j.autcon.2015.12.024}, journal={Automation in Construction}, author={Wang, Dongyun and Zheng, Lijuan and Yu, Hongxiang and Zhou, Wu and Shao, Liping}, year={2016}, month=apr, pages={1–6}, language={en} }

@article{Dao_Na_Nguyen_Ahn_2021, title={\href{https://doi.org/10.1016/j.autcon.2021.103845}{High accuracy contouring control of an excavator for surface flattening tasks based on extended state observer and task coordinate frame approach}}, volume={130}, ISSN={0926-5805}, DOI={10.1016/j.autcon.2021.103845}, abstractNote={In construction, motion control is primary for excavators to complete earth-moving tasks. However, the position tracking performance is strongly affected by system nonlinearity, external disturbances, and model uncertainties during operation. In this paper, a task coordinate frame approach is firstly adopted for an excavator to separate the tracking error into contouring error, tangential error, and orientation error. Based on this transformation, each error component is treated independently according to their priorities. Furthermore, an extended state observer is designed to cope with not only unmeasurable velocities but also lumped disturbances and uncertainties. Finally, these advanced techniques are integrated into the proposed controller by using the backstepping control with the barrier Lyapunov function which is developed to achieve a prescribed performance of the contouring error. The proposed control algorithm guarantees system stability and provides high accuracy contouring performance and acceptable tangential and orientation performances regardless of the presence of lumped disturbances/uncertainties and nonlinearities in the system. Simulation results verify the control effectiveness of the proposed control algorithm in surface flattening tasks compared to previous works. Practitioners can apply the results of the research to not only semi-autonomous operations with unskilled operators but also fully autonomous operations. Future research is necessary to consider the contouring control of excavators with other earth-moving tasks and relating problems in real operating conditions.}, journal={Automation in Construction}, author={Dao, Hoang Vu and Na, Seonjun and Nguyen, Duc Giap and Ahn, Kyoung Kwan}, year={2021}, month=oct, pages={103845} }

@article{Dao_Ahn_2022, title={\href{https://doi.org/10.3390/app12157453}{Active Disturbance Rejection Contouring Control of Robotic Excavators with Output Constraints and Sliding Mode Observer}}, volume={12}, rights={http://creativecommons.org/licenses/by/3.0/}, ISSN={2076-3417}, DOI={10.3390/app12157453}, abstractNote={This paper proposes an active disturbance rejection contouring control scheme for robotic excavators suffering from model uncertainties, external disturbances, and unmeasurable states. A sliding mode observer (SMO) is firstly designed to precisely estimate both joint velocities and lumped uncertainties and disturbances. These estimations are then fed back into the main controller which is constructed based on the task coordinate frame (TCF) approach. Furthermore, to meet the requirements of high-accuracy control performance, the barrier Lyapunov function (BLF) is utilized in the control design together with the previous techniques, which guarantees the stability of the whole system. Finally, numerical simulation is conducted with a high-reliability excavator model to verify the effectiveness of the proposed control algorithm under various operating conditions. In future work, further practical problems will be conducted to realize the application of robotic excavators in construction.}, number={15}, journal={Applied Sciences}, publisher={Multidisciplinary Digital Publishing Institute}, author={Dao, Hoang Vu and Ahn, Kyoung Kwan}, year={2022},  pages={7453}, language={en} }

@article{Yang_Zhang_Hong_Chen_Yang_Wang_Cao_2022, title={\href{https://doi.org/10.1016/j.autcon.2022.104402}{Motion control for earth excavation robot based on force pre-load and cross-coupling compensation}}, volume={141}, ISSN={09265805}, DOI={10.1016/j.autcon.2022.104402}, abstractNote={In the earthmoving automation procedure, the properties of large inertia, strong nonlinearity and flow coupling of multiple actuators are not considered comprehensively in relevant studies, which restricts the accuracy and efficiency performance of leveling tasks. To solve the problem of limited accuracy caused by the difficulty in modeling hydraulic flow coupling and the hysteresis of terminal compensation, we proposed a coordinated motion control method for an automatic earth excavation robot, which reduced the contour error from the perspective of force preloading and terminal control compensation. Various experiments show that the motion coordination of the proposed controller is significantly better than the PID-based controller and the contourbased controller. Due to its stability in multiple conditions, we believe the system will provide an effective earthmoving automation solution for different types of excavators in control. For further study, the soil force estimation will be integrated to improve the system performance under more tasks.}, journal={Automation in Construction}, author={Yang, Teng and Zhang, Bin and Hong, Haocen and Chen, Yuanlong and Yang, Huayong and Wang, Tongman and Cao, Donghui}, year={2022}, pages={104402}, language={en} }

@article{Kim_Kim_Kim_Lee_2019, title={\href{https://doi.org/10.1016/j.automatica.2019.02.041}{Modeling and velocity-field control of autonomous excavator with main control valve}}, volume={104}, ISSN={0005-1098}, DOI={10.1016/j.automatica.2019.02.041}, abstractNote={We propose a novel modeling and control framework for the autonomous excavator with main control valve (MCV), which distributes fluid from pumps to hydraulic actuators with the number of the pumps less than that of the actuators and whose internal hydraulic circuitry switches depending on operating conditions and internal pressures. We first derive the mathematical model of the MCV, including the switching components and supply pump flow constraint. We then design a novel velocity-field control for the bucket position/orientation, which, by relying on a constrained-optimization formulation, can adjust the velocity-field following speed reflecting the physical constraints imposed by the MCV in such a way that the bucket fully follows the desired velocity-field when the constraints are inactive or still preserves the desired direction (or automatic stopping) while slowing down when the constraints become active (e.g. flow saturation). We further show that this optimization can be reduced to simple real-time solvable formulation with its solution existence/optimality (or suboptimality) guaranteed. Simulation is also performed to verify the theory by using a detailed Simulink/Sim-Hydraulics model.}, journal={Automatica}, author={Kim, Kwangmin and Kim, Minji and Kim, Dongmok and Lee, Dongjun}, year={2019},  pages={67–81} }

@article{Wang_Zhang_Hao_Deng_2023, title={\href{https://www.mdpi.com/2227-9717/11/7/1918}{Observer-Based Approximate Affine Nonlinear Model Predictive Controller for Hydraulic Robotic Excavators with Constraints}}, volume={11}, rights={http://creativecommons.org/licenses/by/3.0/}, ISSN={2227-9717}, DOI={10.3390/pr11071918}, abstractNote={Given the highly nonlinear and strongly constrained nature of the electro-hydraulic system, we proposed an observer-based approximate nonlinear model ...}, note={Multidisciplinary Digital Publishing Institute}, number={7}, journal={Processes}, author={Wang, Jian and Zhang, Hao and Hao, Peng and Deng, Hua}, year={2023}, language={en} }

@inproceedings{Msaad_Cecchin_Demir_Fagiano_2025, title={\href{https://ieeexplore.ieee.org/document/11107493}{Data-Driven Model Predictive Control of an Hydraulic Excavator via Local Model Networks}}, ISSN={2378-5861}, DOI={10.23919/ACC63710.2025.11107493}, abstractNote={A novel solution to control an hydraulic excavator during grading tasks is proposed, featuring a Model Predictive Controller designed using Local Model Networks (LMNs), i.e. linear time-invariant dynamic models averaged by nonlinear static functions. The Local Linear Models Tree (LoLiMoT) algorithm is employed to derive an LMN from experimental data of a real excavator. Then, a nonlinear MPC law is designed and implemented on the excavator’s embedded control system. To further improve the computational efficiency, a time-varying MPC law is designed as well, where the LMN is linearized in real-time around the current operating point. Experimental results, conducted with the excavator in real-world conditions, show the effectiveness of both approaches in achieving performance comparable to state-of-the-art solutions, while utilizing a more compact dataset and without the need of the hydraulic cylinders’ pressure measurement.}, booktitle={2025 American Control Conference (ACC)}, author={Msaad, Salim and Cecchin, Leonardo and Demir, Ozan and Fagiano, Lorenzo}, year={2025},  pages={85–90} }

@article{Jin_Ye_Zhang_2023, title={\href{http://arxiv.org/abs/2303.16427}{Learning Excavation of Rigid Objects with Offline Reinforcement Learning}}, DOI={10.48550/arXiv.2303.16427}, abstractNote={Autonomous excavation is a challenging task. The unknown contact dynamics between the excavator bucket and the terrain could easily result in large contact forces and jamming problems during excavation. Traditional model-based methods struggle to handle such problems due to complex dynamic modeling. In this paper, we formulate the excavation skills with three novel manipulation primitives. We propose to learn the manipulation primitives with ofﬂine reinforcement learning (RL) to avoid large amounts of online robot interactions. The proposed method can learn efﬁcient penetration skills from sub-optimal demonstrations, which contain subtrajectories that can be “stitched” together to formulate an optimal trajectory without causing jamming. We evaluate the proposed method with extensive experiments on excavating a variety of rigid objects and demonstrate that the learned policy outperforms the demonstrations. We also show that the learned policy can quickly adapt to unseen and challenging fragmented rocks with online ﬁne-tuning.}, publisher={arXiv}, author={Jin, Shiyu and Ye, Zhixian and Zhang, Liangjun}, year={2023}, language={en} }

@article{Gruetter_Terenzi_Egli_Hutter_2025, title={\href{https://doi.org/10.48550/arXiv.2509.17683}{Towards Learning Boulder Excavation with Hydraulic Excavators}}, DOI={10.48550/arXiv.2509.17683}, abstractNote={Construction sites frequently require removing large rocks before excavation or grading can proceed. Human operators typically extract these boulders using only standard digging buckets, avoiding time-consuming tool changes to specialized grippers. This task demands manipulating irregular objects with unknown geometries in harsh outdoor environments where dust, variable lighting, and occlusions hinder perception. The excavator must adapt to varying soil resistance—dragging along hard-packed surfaces or penetrating soft ground—while coordinating multiple hydraulic joints to secure rocks using a shovel. Current autonomous excavation focuses on continuous media (soil, gravel) or uses specialized grippers with detailed geometric planning for discrete objects. These approaches either cannot handle large irregular rocks or require impractical tool changes that interrupt workflow. We train a reinforcement learning policy in simulation using rigid-body dynamics and analytical soil models. The policy processes sparse LiDAR points (just 20 per rock) from vision-based segmentation and proprioceptive feedback to control standard excavator buckets. The learned agent discovers different strategies based on soil resistance: dragging along the surface in hard soil and penetrating directly in soft conditions. Field tests on a 12ton excavator achieved 70% success across varied rocks (0.4–0.7m) and soil types, compared to 83% for human operators. This demonstrates that standard construction equipment can learn complex manipulation despite sparse perception and challenging outdoor conditions.}, publisher={arXiv}, author={Gruetter, Jonas and Terenzi, Lorenzo and Egli, Pascal and Hutter, Marco}, year={2025}, month=sept, language={en} }

@article{Lu_Zhu_Zhang_2022, title={\href{https://doi.org/10.1109/LRA.2022.3150511}{Excavation Reinforcement Learning Using Geometric Representation}}, volume={7}, ISSN={2377-3766}, DOI={10.1109/LRA.2022.3150511}, abstractNote={Excavation of irregular rigid objects in clutter, such as fragmented rocks and wood blocks, is very challenging due to their complex interaction dynamics and highly variable geometries. In this paper, we adopt reinforcement learning (RL) to tackle this challenge and learn policies to plan for a sequence of excavation trajectories for irregular rigid objects, given point clouds of excavation scenes.Moreover, we separately learn a compact representation of the point cloud on geometric tasks that do not require human labeling. We show that using the representation reduces training time for RL, while achieving similar asymptotic performance compare to an end-to-end RL algorithm. When using a policy trained in simulation directly on a real scene, we show that the policy trained with the representation outperforms end-to-end RL. To our best knowledge, this letter presents the first application of RL to plan a sequence of excavation trajectories of irregular rigid objects in clutter.}, number={2}, journal={IEEE Robotics and Automation Letters}, author={Lu, Qingkai and Zhu, Yifan and Zhang, Liangjun}, year={2022}, month=apr, pages={4472–4479} }

@article{Kim_Kim_2025, title={\href{https://doi.org/10.1007/s12555-024-0213-9}{Enhanced Hydraulic Excavator Control via Semi-automatic Grading Control Using Reinforcement Learning}}, volume={23}, ISSN={2005-4092}, DOI={10.1007/s12555-024-0213-9}, abstractNote={This paper introduces a novel control approach for automated excavators combining model-based and learning-based techniques to enhance control accuracy. The feedback linearization technique is employed based on error dynamics in designing boom and bucket velocity controllers incorporating the driver’s manual arm control. Additionally, supervised learning is used to approximate inverse hydraulic actuation system and to compute joystick control inputs corresponding to the desired control velocity. To further refine control precision reinforcement learning is used to optimize the driver’s manual arm manipulation within a given cycle time. The performance of the proposed methodology is demonstrated through simulations on a 30-ton excavator and compared with results based on model-based techniques.}, number={3}, journal={International Journal of Control, Automation and Systems}, author={Kim, Youngbum and Kim, Jinwhan}, year={2025}, pages={896–906}, language={en} }

@article{Zhang_Zhao_Long_Wang_Qian_Lu_Song_Manocha_2021, title={\href{https://doi.org/10.1126/scirobotics.abc3164}{An autonomous excavator system for material loading tasks}}, volume={6}, DOI={10.1126/scirobotics.abc3164}, abstractNote={Excavators are widely used for material handling applications in unstructured environments, including mining and construction. Operating excavators in a real-world environment can be challenging due to extreme conditions—such as rock sliding, ground collapse, or excessive dust—and can result in fatalities and injuries. Here, we present an autonomous excavator system (AES) for material loading tasks. Our system can handle different environments and uses an architecture that combines perception and planning. We fuse multimodal perception sensors, including LiDAR and cameras, along with advanced image enhancement, material and texture classification, and object detection algorithms. We also present hierarchical task and motion planning algorithms that combine learning-based techniques with optimization-based methods and are tightly integrated with the perception modules and the controller modules. We have evaluated AES performance on compact and standard excavators in many complex indoor and outdoor scenarios corresponding to material loading into dump trucks, waste material handling, rock capturing, pile removal, and trenching tasks. We demonstrate that our architecture improves the efficiency and autonomously handles different scenarios. AES has been deployed for real-world operations for long periods and can operate robustly in challenging scenarios. AES achieves 24 hours per intervention, i.e., the system can continuously operate for 24 hours without any human intervention. Moreover, the amount of material handled by AES per hour is closely equivalent to an experienced human operator.}, number={55}, journal={Science Robotics}, publisher={American Association for the Advancement of Science}, author={Zhang, Liangjun and Zhao, Jinxin and Long, Pinxin and Wang, Liyang and Qian, Lingfeng and Lu, Feixiang and Song, Xibin and Manocha, Dinesh}, year={2021}, month=june, pages={eabc3164} }

@INPROCEEDINGS{10802743,
  author={Werner, Lennart and Nan, Fang and Eyschen, Pol and Spinelli, Filippo A. and Yang, Hongyi and Hutter, Marco},
  booktitle={2024 IEEE/RSJ International Conference on Intelligent Robots and Systems (IROS)}, 
  title={\href{https://www.doi.org/10.1109/IROS58592.2024.10802743}{Dynamic Throwing with Robotic Material Handling Machines}}, 
  year={2024},
  volume={},
  number={},
  pages={98-104},
  doi={10.1109/IROS58592.2024.10802743}}
\clearpage
\appendices
\section{Hydraulic Background}
\FloatBarrier
\subsection{Load Sensing Hydraulic Model}
\label{apx:lsHyd}
The simplified load-sensing circuit in Figure~\ref{fig:ls_hydraulic} shows the pressure feedback loop of an \ac{LS} function.
The pump senses the pressure drop $\Delta P$ across the flow-controlling \ac{DCV}.
It then adjusts pump displacement to keep this pressure drop approximately constant.
As a result, the flow through the valve is mainly set by valve opening and is largely independent of the load pressure at the actuator~\cite{Wang_Wang_2014}.
For control, this makes the hydraulic response comparatively stiff: a given valve command produces nearly the same cylinder velocity over a wide load range.
Our calibrated hydraulic model can therefore map the required flow, or cylinder velocity, directly to a valve command.

\begin{figure}
    \centering
    \includegraphics[width=1.0\linewidth]{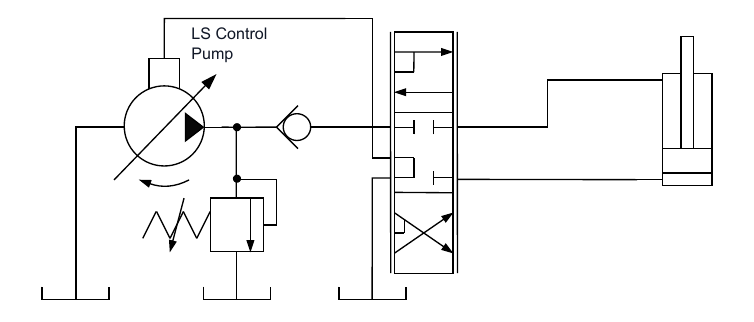}
    \caption{Simplified schematic of a \ac{LS} hydraulic function. The pump is regulated by the function pressure.}
    \label{fig:ls_hydraulic}
\end{figure}

\subsection{Negative Flow Control Hydraulic Model}
\label{apx:nfcHyd}

\begin{figure}
    \centering
    \includegraphics[width=1.0\linewidth]{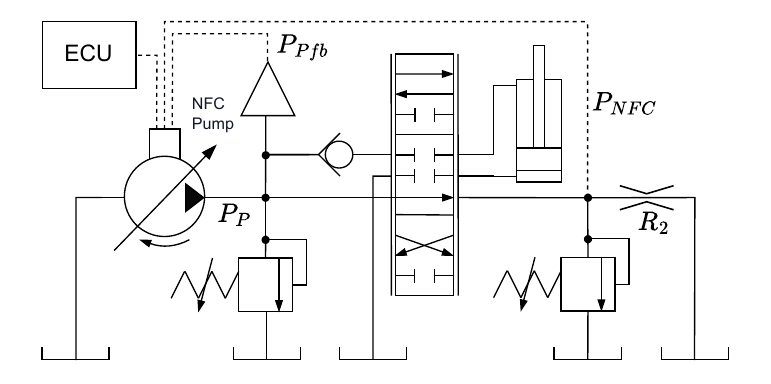}
    \caption{Simplified schematic of a single hydraulic function in an \ac{NFC} circuit. The central bypass is used as a reference for pump control. Pump control remains mainly unaffected by changing load on the function side.}
    \label{fig:nfc_hydraulic}
\end{figure}
Figure~\ref{fig:nfc_hydraulic} shows a simplified \ac{NFC} schematic for a single function.
Since the active hydraulic circuit depends on the travel direction, we model and identify each direction separately. 
We break the hydraulic model down further into functional elements for modeling a single direction in Figure~\ref{fig:nfc_hydraulic_control}.
The abstraction intentionally simplifies the underlying physical effects.
It does not represent all aspects of the true oil flow, but the calculated values match observed and measurable behavior on hardware.
In Section~\ref{sec:experiments}, we provide the experimental validation of the presented model.
For a more comprehensive explanation of \ac{NFC} hydraulics, we refer the reader to \citet{hydbook}, which gives a detailed description of all mechanisms and their differences across implementations, including Caterpillar, Sumitomo, and Kobelco.

We decompose the \ac{DCV} into two active orifices, one for the \ac{NFC} central bypass $R_1$ and one for the function power line $R_3$. 
The resistances of $R_1$ and $R_3$ depend on spool displacement and are inversely proportional to each other. 
We further assume $R_2 \gg R_1$, so the \ac{NFC} path flow $Q_{\text{NFC}}$ is negligible.
\ac{NFC} uses a variable-displacement pump whose stroke decreases with increasing control-line pressure $P_{\text{NFC}}$, that is generated by the accumulative resistance of the central bypass $R_1$ with a reference orifice $R_2$. 
Opening the \ac{DCV} reduces the bypass-orifice-induced pressure and increases pump flow. 
An additional feedback line from pump pressure $P_P$ and the \ac{ECU} can destroke the pump as well. 
Since this core dynamic is load independent, the system does not react \textit{stiff} to valve commands, as described in greater detail by~\cite{insaneHydraulics}.
In our model, the flow into the function is determined by the balance between the back pressure $P_{\text{func}}$, the function orifice $R_3$, and the pump pressure $P_P$, that is determined by the bypass galleries $R_1$, and the reference orifice $R_2$. 
The bypass galleries and reference orifice together with the \ac{NFC} pump and $P_{Pfb}$ form a pressure control loop that depends only on the \ac{DCV} displacement.
We can probe this mapping by stalling the function ($Q_\text{function}$ = 0) and measuring the function pressure $P_{func} = P_P$.
Only when $P_P$ exceeds the cylinder back pressure can oil flow into the function. 
The resulting flow into the function depends only on the pressure difference $P_p - P_{\text{func}}$ and the resistance of orifice $R_3$. 
For calibration, we characterize the pressure-control feedback loop as a feed-forward function and the orifice function of $R_3$.

\begin{figure}
    \centering
    \includegraphics[width=\linewidth]{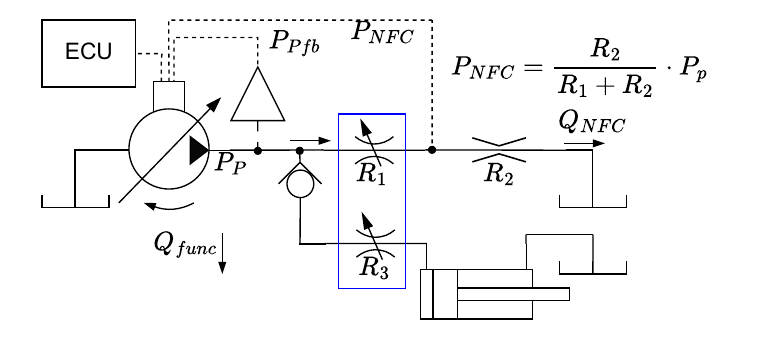}
    \caption{Broken down schematic of \ac{NFC} for modeling. Highlighted in blue are the abstracted orifices from the \ac{DCV} in an actuated position. The flow resistance follows a curve based on the spool displacement of the valve.}
    \label{fig:nfc_hydraulic_control}
\end{figure}

\section{Robots}
Table \ref{tab:excavator_comparison} compares the two platforms used for evaluation.
The Menzi Muck M445 as shown in Figure~\ref{fig:menzi} is controlled through the vendor remote control interface.
A cabin mounted computer is running our controller and sends valve commands to the excavator through CAN-Bus.
Inertial, joint position and joint velocity data is measured by a Leica Icon Machine Control system.

CASE250 in Figure~\ref{fig:case} is controlled through a Husco Exacto machine control system.
Similarly to M445, a cabin mounted computer receives joint measurements from a Leica Icon Machine Control system, runs the controller presented in this paper, and sends valve commands to the Exacto.

\begin{table}[ht]
\centering
\caption{Comparison of Menzi Muck M445 and CASE250 Excavators}
\label{tab:excavator_comparison}
\begin{tabular}{|l|c|c|}
\hline
\textbf{Parameter} & \textbf{M445} & \textbf{CASE250} \\
\hline
Weight                & 11.5\,t          & 25\,t        \\
Engine Power          & 105\,kW          & 138\,kW      \\
Overall Length        & 6.1\,m           & 5.3\,m       \\
Overall Width         & 2.3\,m           & 3.2\,m       \\
Boom Length           & 2.5\,m           & 5.8\,m       \\
Stick Length          & 1.9\,m           & 3.0\,m       \\
Bucket Length         & 1.3\,m           & 1.6\,m       \\
Bucket Capacity       & 0.54\,m$^3$      & 0.8\,m$^3$   \\
Maximum Reach         & 7.8\,m           & 10.8\,m \\
\hline
\end{tabular}
\end{table}

\begin{figure}
    \centering
    \includegraphics[width=1.0\linewidth]{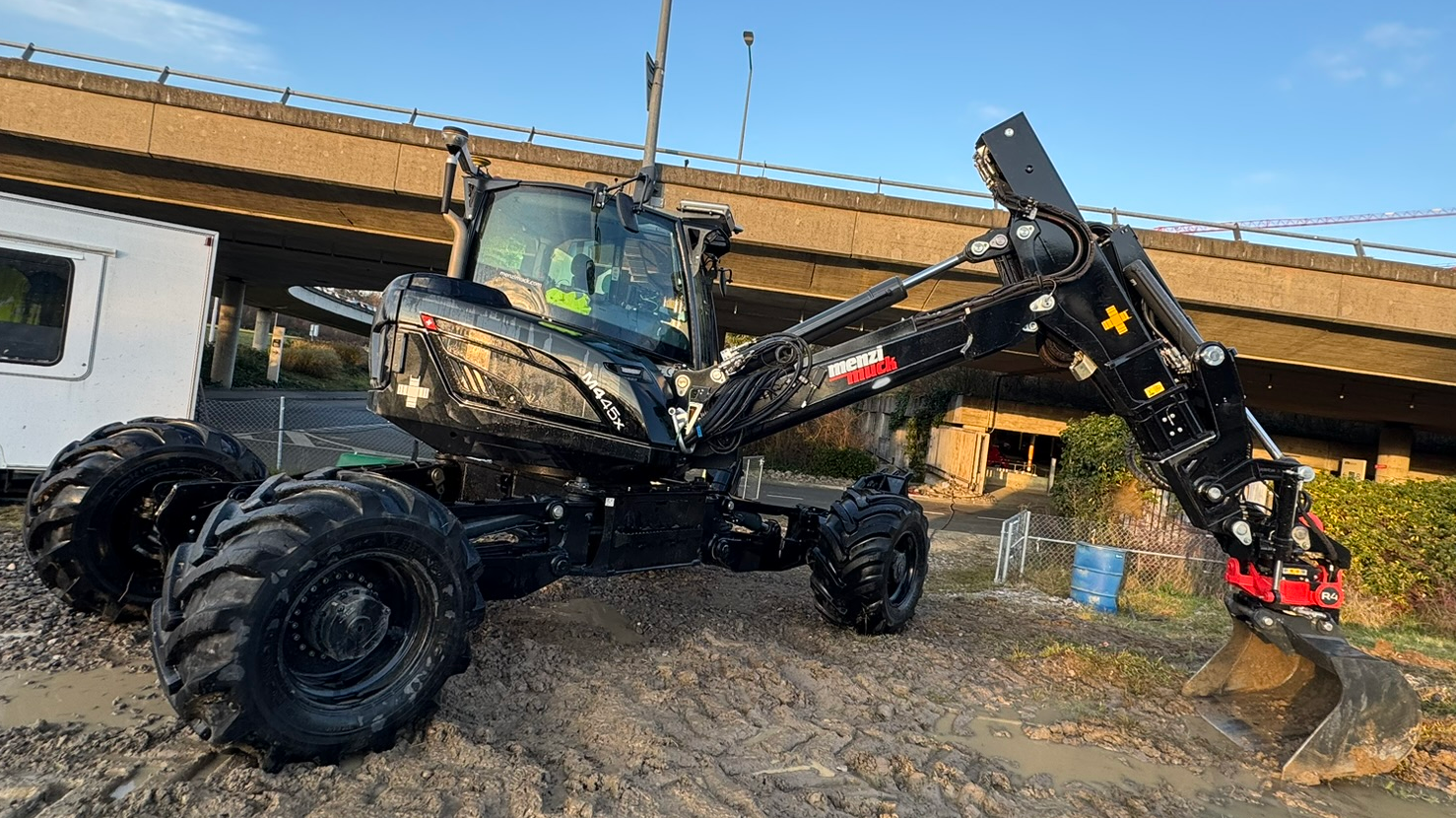}
    \vspace{-10pt}
    \caption{Menzi Muck M445}
    \label{fig:menzi}
\end{figure}

\begin{figure}
    \centering
    \includegraphics[width=1.0\linewidth]{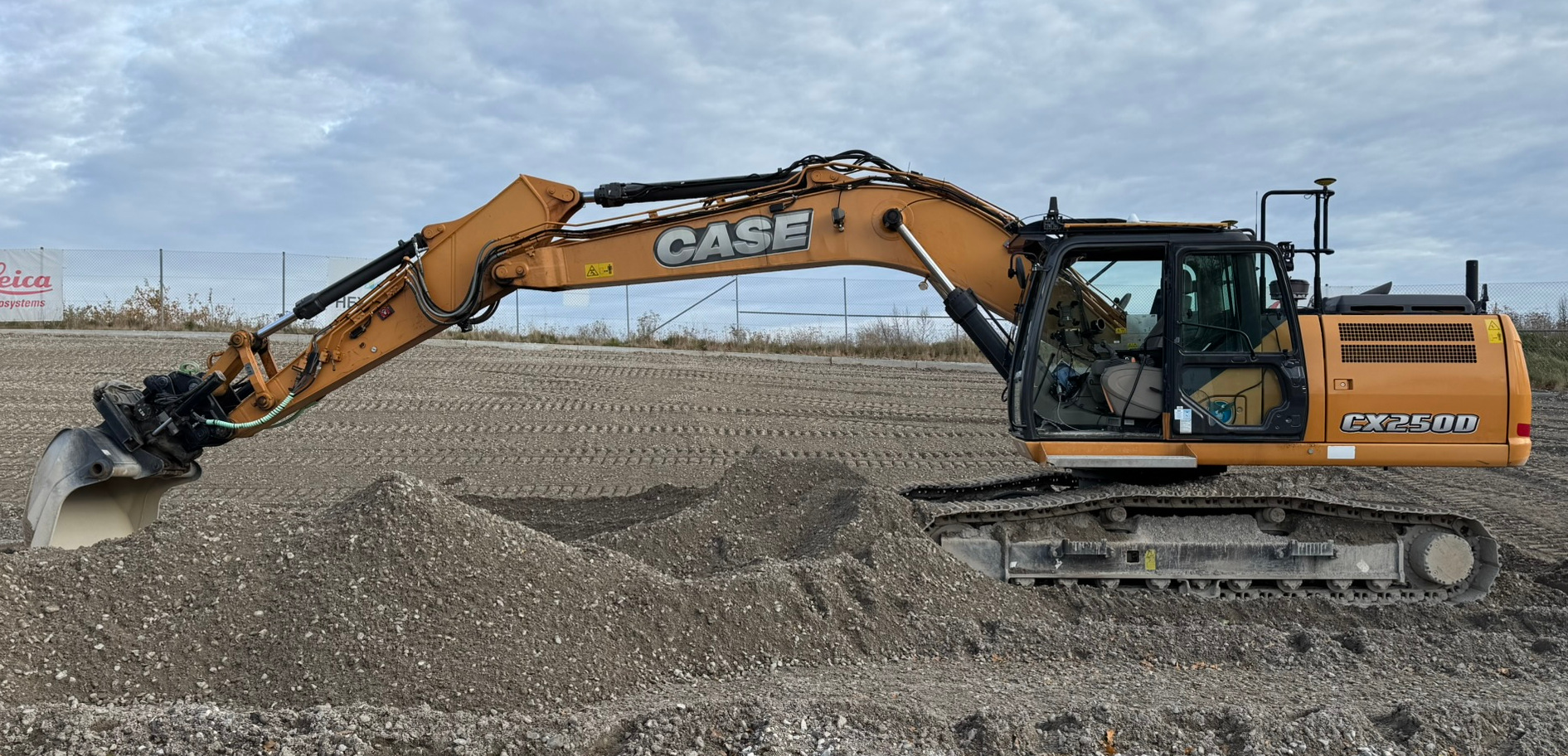}
    \vspace{-10pt}
    \caption{CASE250}
    \label{fig:case}
\end{figure}

\end{document}